\documentclass{article}
\usepackage[square,numbers]{natbib}

    \usepackage[final]{neurips_2023}

\usepackage[utf8]{inputenc} % allow utf-8 input
\usepackage[T1]{fontenc}    % use 8-bit T1 fonts
\usepackage{hyperref}       % hyperlinks
\usepackage{url}            % simple URL typesetting
\usepackage{booktabs}       % professional-quality tables
\usepackage{amsfonts}       % blackboard math symbols
\usepackage{nicefrac}       % compact symbols for 1/2, etc.
\usepackage{microtype}      % microtypography
\usepackage{xcolor}         % colors
\usepackage{subfigure}
\usepackage{enumitem}
\usepackage{todonotes}
\usepackage{amsmath}
\usepackage[title]{appendix}

\title{Distributed Reinforcement Learning for Molecular Design: Antioxidant case}

\usepackage{authblk}
\author[1]{Huanyi Qin}
\author[2]{Denis Akhiyarov}
\author[2]{Sophie Loehle}
\author[1]{Kenneth Chiu}
\author[2]{Mauricio Araya-Polo}
\affil[1]{Department of Computer Science, SUNY Binghamton\hfil \{hqin4, kchiu\}@binghamton.edu}
\affil[2]{TotalEnergies\hfil \{denis.akhiyarov, sophie.loehle, mauricio.araya\}@totalenergies.com}
\begin{document}

\maketitle

\begin{abstract}
Deep reinforcement learning has successfully been applied for molecular discovery as shown by the Molecule Deep Q-network (MolDQN) algorithm. This algorithm has challenges when applied to optimizing new molecules: 
training such a model is limited in terms of scalability to larger datasets and the trained model cannot be generalized to different molecules in the same dataset.
In this paper, a distributed reinforcement learning algorithm for antioxidants, called DA-MolDQN is proposed to address these problems.
State-of-the-art bond dissociation energy (BDE) and ionization potential (IP) predictors are integrated into DA-MolDQN, 
which are critical chemical properties while optimizing antioxidants.
Training time is reduced by algorithmic improvements for molecular modifications. The algorithm is distributed, scalable for up to 512 molecules, and generalizes the model to a diverse set of molecules.
The proposed models are trained with a proprietary antioxidant dataset. The results have been reproduced with both proprietary and public datasets. The proposed molecules have been validated with DFT simulations and a subset of them confirmed in public "unseen" datasets.
In summary, DA-MolDQN is up to 100x faster than previous algorithms and can discover new optimized molecules from proprietary and public antioxidants.
\end{abstract}
% In a word, DA-MolDQN is 2.6x faster than MT-MolDQN and can optimize both proprietary and public antioxidants.

\section{Introduction}
% to do: explain more about why 
Antioxidants are compounds that inhibit oxidation and are critical in a variety of industrial applications,
such as fuel additives \cite{anti_application_1}, lubricants \cite{anti_application_2}, and polymer stabilization \cite{anti_application_3}. 
They can prevent oxidation and degradation, improving the performance and longevity of materials and products.
Traditional antioxidant discovery approaches are time-consuming and costly,
%\todo[inline]{add some references.}
%
because these approaches typically require time-consuming iterative approaches from chemical researchers and rely on expensive Density Functional Theory (DFT) \cite{parr1979local} calculations.
Reinforcement learning (RL) algorithms \cite{rl_mol_1,rl_mol_2,rl_mol_3,rl_mol_4} are often used 
to optimize and generate new molecules based on their chemical properties,
such as quantitative estimation of drug-likeness (QED) or synthesizability (SA) scores.

However, new challenges arise while using RL to optimize antioxidants.
Bond dissociation energy (BDE) and ionization potential (IP) indicate the reactivity and stability of the antioxidant molecules, respectively \cite{anti_bde1, anti_bde_ip}.
Unfortunately, the estimation of BDE and IP is more demanding than QED and SA scores because
DFT calculation is borderline prohibitive---
% , which is not the case for QED and SA scores.
% Traditionally, calculating BDEs using the DFT method can be computationally demanding, taking hours or even days for a single BDE value \cite{bde_dft}. 
% Traditionally, calculating BDEs using the DFT method can be computationally demanding, 
it takes hours or even days to calculate a single BDE value while using DFT \cite{bde_dft}. 
%
% It takes several hours or even several days to process a DFT simulation, while the computation time of QED and SA scores is less than 1 second.
In this paper, Alfabet \cite{alfabet} and AIMNet-NSE \cite{AIMNet-nse} machine learning models are integrated to predict and estimate the BDE and IP properties of antioxidants, 
which are state-of-the-art BDE and IP predictors based on deep learning.
%
% Compared to traditional methods, the property predictors achieve high accuracy and are much faster.
%
% Their performance is further improved by introducing a Least Recently Used (LRU) cache \cite{LRU_cache} and reducing the needed 3D conformers.

% The performance of BDE predictor is further improved by introducing a Least Recently Used (LRU) cache \cite{LRU_cache}.
% %
% In order to speed up IP predictor, the needed 3D conformers are also reduced without losing too much accuracy.

Another challenge of antioxidant properties optimization is that the Pareto optimization front between BDE and IP is often a trade-off.
Molecules with good BDE properties usually have poor IP properties and vice versa.
% to do: add Pearson product-moment correlation coefficient of BDE and IP
%
% Molecules with good BDE properties may be more reactive, but these molecules usually have bad IP properties which indicate they are unstable and have no commercial value for practical applications.
%
% Conversely, the molecules with good IP but bad BDE are stable but have weak antioxidant properties.
%
% In a word, the ideal antioxidants must have both excellent BDE and IP properties.
%
By balancing the weight of BDE and IP, the proposed RL agent is able to optimize both BDE and IP properties of antioxidants,
so that the optimized molecules have stronger antioxidant properties and are more stable.

MolDQN \cite{moldqn} algorithm is a well-known RL model for optimizing molecules.
%
% Subsequently, multi-task (MT)-MolDQN was developed \cite{mt-moldqn}, with a major advantage over the original MolDQN algorithm--it is able to train the model with several initial molecules simultaneously. 
Subsequently, multi-task (MT)-MolDQN was developed \cite{mt-moldqn}, which has a major advantage over the original MolDQN algorithm in that it is able to train the model with multiple initial molecules simultaneously. 
Their optimizations start with one or several initial molecules, 
but their explorations are restricted to the neighborhood of initial molecules,
which prevents them from exploring larger antioxidant datasets and proposing well-optimized antioxidants.
To address this, the distributed DA-MolDQN was proposed, which is capable of training and optimizing hundreds of antioxidants simultaneously. 
% As shown in Figure~\ref{fig:random_walk}, the MolDQN and MT-MolDQN are far from exploring the antioxidant universe, so their proposed antioxidants are not well optimized, especially for unseen molecules.
% To address this, the distributed DA-MolDQN was proposed, which is capable of training and optimizing hundreds of antioxidants simultaneously. 
% The general model is trained by DA-MolDQN on the proprietary antioxidant data set and it is able to optimize molecules for both proprietary and public antioxidants testing data sets.

In this paper, extra methods and optimizations are also necessary to 
% integrate the predictors, optimize antioxidants, and 
further improve the performance and efficiency. 
They include:
(1) additional protection mechanism of O-H bond,
(2) an efficient method that avoids invalid 3D conformers for new molecules,
(3) improved C++ RL environment ported from original Python implementation,
(4) incremental Morgan fingerprint algorithm,
(5) a BDE property cache,
(6) a filtering algorithm to constrain the search space,
% (7) the fine-tuned model for optimizing outlier molecules.
(7) and the optional fine-tuned steps for further optimizing outlier molecules.

Specifically,
we make the following contributions:
% \begin{itemize}
%     \item 
%     DA-MolDQN integrates state-of-the-art BDE and IP predictors. 
%     \item 
%     The molecules optimized by DA-MolDQN have both good BDE and IP, which have strong antioxidant properties and are stable.
%     \item 
%     The model trained by DA-MolDQN is generalizable and can optimize antioxidants on both proprietary and public data sets.
%     \item 
%     DA-MolDQN is efficient and is 2.6x faster than MT-MolDQN, which is important for training a general model. And the performance of predictors is also improved.
% \end{itemize}
\begin{itemize}
    \item 
    DA-MolDQN integrates state-of-the-art BDE and IP predictors. 
    \item 
    The molecules optimized by DA-MolDQN have both good BDE and IP, which have strong antioxidant properties and are stable.
    \item 
    The model trained by DA-MolDQN is generalized and can optimize antioxidants on both proprietary and public data sets.
    \item 
    DA-MolDQN is efficient and is 2.6 - 106x faster than MT-MolDQN and MolDQN.
\end{itemize}

% The following sections first introduce the predictors and RL algorithms. 
% Then it presents the details of the proposed distributed algorithm. 
% The results of optimization experiments and their DFT validation are shown in the next sections. 
% The conclusion and future work are also discussed at the last of the paper.

\section{Background \& Related Work}\label{sec:background_and_related_work}

\subsection{Chemical Properties of Antioxidants}\label{sec:chemical_property}
% Bond dissociation enthalpy (BDE) and ionization potential (IP) are important chemical properties 
% when looking for new antioxidants because they indicate the reactivity and stability of the molecules, respectively \cite{anti_bde1, anti_bde_ip}.
%
% An ideal antioxidant is expected to have both excellent BDE and IP properties, otherwise, it will be inefficient or unstable.
\paragraph{BDE Property of O-H bond}

BDE is used to measure the strength of a chemical bond.
BDE of the O-H bond is one of the most important characteristics of an antioxidant \cite{bde_ms, bde_r2, anti_bde1, anti_bde_2}. 
In this paper, BDE refers to the lowest BDE values of all O-H bonds in the molecule.
%
% The BDE assesses the potency of a phenolic compound to yield its H atom toward a radical.
%
The lower the BDE, the easier it is for the antioxidant to yield its hydrogen atom, and the more promising the compound is. 
For the antioxidant to be effective, the BDE of the antioxidant must be well below the BDE of peroxide radicals (88-90 kcal/mol) \cite{bde_oh_88_90}. 
Specifically in this work, the BDE of optimized antioxidants should be lower than 76 kcal/mol,
which are thought to have good BDE properties.
\paragraph{IP Property}
% Ionization potential (IP)
IP is the energy required to remove an electron from an atom. 
It tells how tightly the electron is bound and how stable the antioxidant is.
If the IP is too low, the compound becomes unstable in air 
and undergoes natural oxidation with the dioxygen present in the air, losing its antioxidant activity and making it useless for antioxidants.
The IP of generated antioxidants should be higher than 145 kcal/mol.

\paragraph{Trade-off between BDE and IP}\label{para:trade_off_between_bde_and_ip}
The BDE can be efficiently lowered by incorporating electron-donor substituents on the ring such as methyl, methoxy, tertiobutyl, or hydroxide, preferably in ortho and para position.
It allows stabilization of the formed radical, leading to a lower BDE. 
However, it’s not possible to stack five dimethyl amino groups on the ring and hope to find the best antioxidant ever made. 
This doesn’t work because introducing electron-donating substituents lowers both the BDE and the IP \cite{anti_bde1}.
As mentioned above, when IP is too low, the compound becomes unstable and useless. 
There is a trade-off between BDE and IP, and the RL agent needs to balance the BDE and IP so that the generated antioxidants are effective and stable.
%
% To address this, a reward function is introduced to combine the normalized BDE and IP. 
% %
% The reward function helps the RL agent to optimize the molecule's properties by considering the influence of modifications on both BDE and IP. 
% %
% And the RL agent can generate effective and stable antioxidants with improved properties.

% \paragraph{Other Expected Properties of Antioxidants}

\subsection{State-of-the-art Property Predictors}
\paragraph{BDE Predictor: Alfabet}
% Alfabet is based on machine learning algorithms and has been shown to be accurate in predicting BDE for a wide range of molecules. 
%
% Alfabet leverages the power of machine learning, particularly graph neural networks (GNNs), to accelerate the prediction of BDEs. 
%
% The GNNs learn from graph-based molecular representations, allowing for more efficient and accurate BDE predictions.
Alfabet is based on graph neural networks (GNNs) and has been shown to be accurate in predicting BDE for a wide range of molecules.
When using Alfabet to predict the BDE of generated molecules, it accepts SMILES representation of molecules as input.
%and returns the BDE of all bonds in molecules. 
%
Then the lowest BDE is found among all O-H bonds and it is termed as BDE in this paper.
%
% The prediction accuracy of Alfabet's pre-trained model is tested on the proprietary antioxidant data set and the predicted BDE of O-H is compared with real BDE (DFT value). 
%
% The average relative error is less than 5\% on the above data set.
%

\paragraph{IP Predictor: AIMNet-NSE}
AIMNet-NSE is also a machine learning architecture that can predict molecular energies including IP,
%
% and it achieves high accuracy for lots of molecules. 
and it achieves high accuracy from training with over 100k molecules. 
%
% Same to BDE prediction, the traditional DFT method is very slow in predicting IP property \cite{ip_dft}, 
% %
% while the AIMNet-NSE introduces a significant prediction performance improvement.
%
% The AIMNet-NSE uses the 3D conformer of molecules to predict IP properties.
%
% In a molecule, the 3D conformer includes the 3D coordinates and atom information of all atoms.
%
% However, the need for 3D conformers brings new challenges to the RL algorithm and the details are explained in 
% Section~\ref{sec:avoiding_invalid_3d_conformers}.
% next Section.
%
% The accuracy of AIMNet-NSE is also measured  on the proprietary antioxidant data set
% and the average relative error is less than 3\%.
The AIMNet-NSE uses the 3D conformer of molecules to predict IP properties, which brings new challenges, 
because the molecules generated by MolDQN algorithm may not have a valid and stable 3D conformer.

Both predictors have high accuracy on the proprietary antioxidant data set. Their average relative error is less than 5\%.
Although their performance is much faster than DFT, they are still the main bottlenecks for RL optimization.
Their performance is further improved by introducing a cache and reducing the needed 3D conformers.
% Least Recently Used (LRU) cache \cite{LRU_cache} 

% to do: some predicted IP is super large and even is 

\subsection{MolDQN \& MT-MolDQN} \label{sec:moldqn}
MolDQN is an innovative deep reinforcement learning model that treats molecules as undirected graphs,  
optimizing the structure of molecules by adding or removing new atoms and bonds.
% The typical modifications include: 
% (1) Atom Addition: add an atom and connect it to an atom in the original molecule,
% (2) Bond Addition: add a new bond to connect two atoms.
% (3) Bond Removal: remove a bond and unconnected atoms.
The modifications restart from the initial molecules in every episode and the molecule is slightly optimized in each step.
So the explorations are around the initial molecules.
Each modification will take care of the valence of different atoms, and the modified molecules are only guaranteed to be valid in 2D graph representations.
The modified molecules may not be valid in 3D space and have valid 3D conformers because these modifications do not consider the 3D structures, such as the torsion angles \cite{torsion_angle} and aromatic ring.
% After each modification, the RL agent will input the modified molecules into property predictors and calculate their rewards.
After each modification, the modified molecules are inputted into property predictors, then their rewards are calculated. 
The molecules and other needed information are stored as samples to train the model.

% The RL agent will then store and learn from the experiences and eventually find a modification path to optimized molecules.
MolDQN has achieved excellent results on both single or multiple objective optimization tasks to find molecules with better or specific properties, such as maximizing penalty logP values and
% quantitative estimation of drug-like properties (QED) values. 
QED.
% By using a DQN approach that introduces self-sampling, the model achieves deep exploration and outperforms the plain DQN model and other state-of-the-art algorithms in specific cases.
% After that, a well-trained model can also be used to optimize unseen molecules.
The following MT-MolDQN is the parallel version and it is implemented with Pytorch Distributed Data-Parallel (DDP) \cite{pytorch_ddp}. 
Compared to MolDQN, MT-MolDQN proposes better-optimized molecules and is more efficient, 
because more initial molecules can be used in training.

\section{Distributed Antioxidant Optimizer: DA-MolDQN}
% \subsection{Distributed and Batched Algorithm}
% The flowchart in Figure~\ref{fig:flowchart} shows the workflow of DA-MolDQN.
% The workflow of DA-MolDQN is shown in Figure~\ref{fig:flowchart}, 
% which describes how the proposed general model is trained.
% The flowchart starts with an overview of distributed processes.
% Then it shows how the molecules are optimized and evaluated within a process.
% Other contributions, such as integrating property predictors, distributed training, reward function, filter script, and fine-tuning, 
% are all present in the flowchart 
% and the details of the workflow will be explained in this section.
% The performance optimizations are also present at the end of the section.

Figure~\ref{fig:flowchart} illustrates the workflow of DA-MolDQN, providing an overview of the training process for the general model.
The flowchart commences with an introduction to distributed processes, followed by a depiction of the optimization and evaluation of molecules within these processes.
The flowchart also includes various other elements, encompassing the integration of property predictors, distributed training, reward function, filter script, and fine-tuning.
This section elaborates on the intricacies of the workflow and also presents performance optimizations.

% Distributed training,  performance optimization is also introduced.

\subsection{Distributed Overview \& Details in Process}
\begin{figure}[ht]
  \includegraphics[trim=0 0 0 0,clip,width=\textwidth]{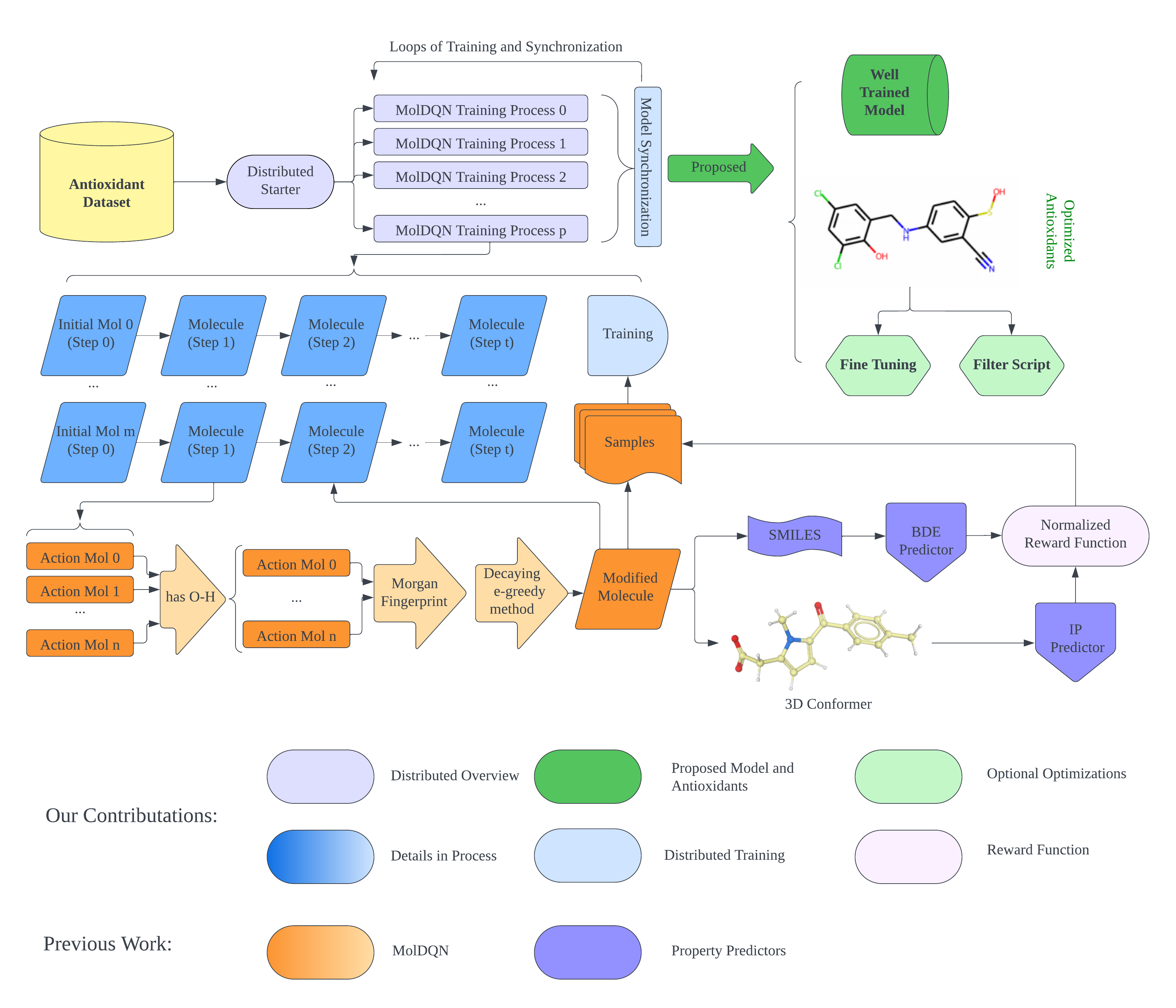}\hfil
% \caption{This figure shows the workflow and DA-MolDQN in detail. 
% It starts with an overview of the whole distributed training process group 
% and further explains the modification of molecules, property prediction and training with samples in each process.
% It then presents the trained models and optimized antioxidants.
% The optional optimizations such as fine-tuning and filter scripts are also included.
% %
% }
\caption{Workflow of DA-MolDQN}
\label{fig:flowchart} 
\end{figure}

\paragraph{Distributed Overview}
The MT-MolDQN uses Pytorch DDP as a framework, and DA-MolDQN extends it to a distributed training script, so the DA-MolDQN is no longer limited by the computational resources within a node. 
The distributed processes (workers) are launched by SLURM \cite{slurm} and each process will receive one initial molecule.
The processes will work on optimizing their initial molecules independently, but they will cooperate to propose a well-trained and generalized model.

\paragraph{Details in Process: Step by Step Optimization}
The molecule is modified and optimized by the RL agent in each process. 
This actually generates an optimization path to the final proposed molecule, and the initial molecule is optimized step by step.
In short, one step of a molecule includes: generating all valid action molecules based on chemical valence, choosing an action by decaying epsilon greedy method \cite{intro_rl}, and predicting the properties.
The details are explained in Section~\ref{sec:moldqn} and the MolDQN paper \cite{moldqn}.

\paragraph{Details in Process: Batched Modification}
Batched modifications (blue parallelograms in Figure~\ref{fig:flowchart}) are needed so that each process can process multiple initial molecules, which in turn promote parallelism. 
Each process receives multiple initial molecules while launching.
In each step, the batched RL algorithm will operate on all molecules in that process one by one.
It will not go to the next step until all molecules in the current step finished their operations.
By using batched modification, the number of initial molecules further increases, and diverse samples could be generated by the RL agent.
The diversity of samples is crucial to train a general model. 
The batched modification also reduces the required computational resources because the multiple molecules in a process share the expensive BDE and IP property predictors. 

% \paragraph{Sample Collection, Training \& Model Synchronization}
\subsection{Distributed Training}
Each process has a replay buffer that stores all modified molecules and their useful information, such as fingerprints \cite{morganfp}, and rewards.
The samples from the replay buffer are reused to train the model.
The losses are collected from all processes during training.
Then, the general model is synchronized among all processes at the end of every episode,
so the experience learned from other processes will be utilized in the next episodes. 

\subsection{Extra Efforts on Integrating Property Predictors}
\paragraph{BDE: Protecting O-H Bond}
Since the BDE property is the lowest BDE among all O-H bonds,
an implicit restriction is that the generated molecules must have at least one O-H bond, 
otherwise, the BDE properties are undefined and the molecules are not needed.
This restriction is not guaranteed in the MolDQN and MT-MolDQN, 
because the O-H bond may be accidentally broken by the modifications.
These modifications are not allowed, 
and the molecules always have at least one O-H bond in this paper.
The approach removes only a few invalid modifications (e.g. Action Mol 1 in Figure~\ref{fig:flowchart}) from over a hundred action molecules, which has a trivial impact on the exploration of molecular space.
The examples are presented in Appendix~\ref{app:protection_of_oh}.

\paragraph{IP: Avoiding Invalid 3D Conformers}\label{sec:avoiding_invalid_3d_conformers} % moved to appendix
AIMNet-NSE uses the 3D conformers of the molecules as its input. %, while the Alfabet uses SMILES strings. 
The RDKit \cite{rdkit} is used to calculate the 3D coordinates of all atoms in the molecules \cite{embedMolecule1}. 
However, as mentioned in Section~\ref{sec:moldqn}, some generated molecules may not have a valid 3D conformer.
The MolDQN has done some work on this, such as limiting the size of new rings, but their efforts are still far from being complete. 
In order to avoid generating molecules without 
any valid 3D conformations, 
the reward of invalid molecules is set to -1000,
which is much less than the normal rewards (0.8-2.5).
Then RL agent successfully learns how to avoid these invalid conformers, without employing a huge number of human handwritten chemistry rules.
An example of an invalid molecule and the results of the avoiding approach are shown in Appendix~\ref{app:avoiding_invalid_3d_conformers}

% \subsection{Normalized Reward \& Reward Function}\label{sec:normalized_reward_and_reward_function}
\subsection{Normalized Reward Function}\label{sec:normalized_reward_and_reward_function}
As mentioned in Section~\ref{sec:background_and_related_work}, 
% As mentioned in previous Section, 
the reward function needs to combine the normalized BDE and IP properties.
In this paper, min-max normalization is used for both BDE and IP. 
The lower bound and upper bound are minimal and maximum properties in the proprietary data set.
The weights of the normalized rewards are also useful to further balance the BDE and IP, depending on the optimization target. 
They are set to 0.8 and 0.2 in this paper because the BDE property is related to the effectiveness of antioxidants and is more important than IP.
Different combinations of reward weights are also tested and the above weights are found to be optimal.
%
% \paragraph{"Simpler is Better"}

The molecules with fewer atoms and bonds are preferred when seeking new antioxidants, so the $\gamma$ is also added to Equation~\ref{Eq:reward}. 
It represents the relatively reduced atoms and bonds from the initial molecule. 

\begin{equation}
\text{Reward} = - w_1 \cdot \text{nBDE} + w_2 \cdot \text{nIP} + w_3 \cdot \gamma
\label{Eq:reward}
\end{equation}

\subsection{Optional Optimizations}
\paragraph{Filter Script}\label{sec:filter_script}
% Although BDE and IP are the most important chemical properties of antioxidants, there are several other properties that chemists are interested in, such as Tanimoto similarity \cite{tanimoto} of fingerprints and SA score.
% Secondly, the proposed model has optimized the antioxidants a lot, but the generated molecules are not 100\% guaranteed to fit the BDE and IP restrictions.
Although the antioxidants are significantly optimized by the general model, the molecules are not 100\% guaranteed to fit the BDE and IP constraints.
Further, there are several other properties that chemists are interested in, such as Tanimoto similarity \cite{tanimoto} of fingerprints and SA score.
To address this, an extra script filters out molecules without good BDE and IP properties.
The molecules are also filtered out if their SA scores are higher than 3.5 or if they are identical to existing antioxidants. 
% and their optimization processes are marked as failed. 
% Tanimoto similarity for the generated molecules to all proprietary antioxidants is also calculated in the filter script. 
% The generated molecules are expected to be similar to the original antioxidants because they are more likely to be good candidates, while the molecules that are identical to existing antioxidants are filtered out.
% There is no hard limit on the SA score now, but the lower the SA score the better for antioxidants. The molecules are also filtered out if their SA scores are higher than 3.5. 

% \paragraph{Fine-Tuning of General Model}
\paragraph{Fine-Tuning}
% The general model may not be applicable to all molecules. 
% Some molecules may require specific models to get better modifications and these modifications are absent in other antioxidant optimizations. 
The experiences that a general model learns from the most antioxidant optimizations may prevent the RL agent from doing some unusual but effective optimizations
for the outlier molecules.
Inspired by \cite{fine_tune_1, fine_tune_2}, a few fine-tuning episodes could significantly improve the performance of a model in special environments.
An optional fine-tuning step is introduced to the workflow.
% The chemical properties and training success of these molecules are significantly improved.
The fine-tuning starts with the pre-trained general model,
and the initial epsilon threshold is 0.5. %in the experiments. % since the models have already acquired some knowledge through general training. 
Fine-tuning is independent for each molecule and the properties of irregular molecules are further improved with trivial overhead.

% The general models can produce better antioxidants when optimizing molecules, but they are not applicable to all molecules. Some molecules may require specific models to get better modifications and these modifications are absent in other antioxidant optimizations. What's more, the experiences that a general model learns from other antioxidant optimizations may even prevent the RL agent from doing some unusual but very powerful optimizations.
% Inspired by those papers \cite{fine_tune_1, fine_tune_2}, a few fine-tuning episodes could significantly improve the performance of a model in special environments.
% The chemical properties and training success of these molecules are significantly improved.
% The fine-tuning process starts with the pre-trained general models. 
% The initial epsilon threshold is 0.5 in the experiments since the models have already acquired some knowledge through general training. 
% Fine-tuning is done independently for each molecule and the properties of irregular molecules can be better optimized.

% Some optimized antioxidants may be worse after fine-tuning. In this case, the user should simply use the result of the general model.
% \paragraph{Trivial Overhead of Fine-Tuning}
% Typically, it takes about 120 seconds to fine-tune a molecule, which is trivial compared to training an initial model from scratch.
% % Because the fine-tuning needs much fewer episodes and has a high LRU cache hit rate.
% Each molecule will have its own model during fine-tuning. But there is no need to store those models after the fine-tuning process.

\subsection{Non-trivial Performance Improvement} % moved to appendix
% \section{Non-trivial Performance Improvement}

\paragraph{General Performance Optimization}
The MolDQN and MT-MolDQN are profiled with py-spy \cite{py-spy} and the result demonstrates that the molecule modification and Morgan fingerprint \cite{morganfp} calculation are the main performance bottlenecks of MT-MolDQN.
To address this, the modification functions are re-implemented in C++ instead of Python, and a fast incremental Morgan fingerprint algorithm is developed.
With the help of the above optimizations, DA-MolDQN is 2.6x faster than MT-MolDQN while using QED and SA score rewards.

\paragraph{Property Predictor Performance Optimization} %to do: Draw a figure to use the LRU cache hit rate
Although Alfabet and AIMNet-NSE are much faster than the traditional DFT method, they are still 466.8x and 32.6x slower than QED calculation. 
% The predictors introduce non-trivial computational overhead and t
The estimated computation time without any performance optimization will be over 16 days.
To address this, a Least Recently Used (LRU) \cite{LRU_cache} cache is introduced to store the predicted BDE values.
The AIMNet-NSE proposed 5 trained models and suggested using the average predicted properties. 
However, only one model is used in this paper and the accuracy is still good enough.
The performance of predictors is significantly improved during training and fine-tuning.

% take about 80\% of the computation time in the training scripts.
%
% To train a model and optimize an antioxidant, the estimated computation time of the initial scripts without any performance optimization will be over 16 days, 
% %
% which are obviously too slow to be commercially and practically applicable.
%
% To address this, a Least Recently Used (LRU) \cite{LRU_cache} cache is introduced to store the action molecules that are generated in the training and their predicted BDE values.
% %
% The overhead of LRU cache is trivial compared to the BDE predictor, and the cache hit rate increases greatly during the training. 
% %
% Because the epsilon rate of the DQN algorithm decreases and the RL agent prefers greedy actions more and more. 
% %
% The LRU cache significantly improved the training and fine-tuning performance. 
% %
% The AIMNet-NSE proposed 5 trained models and suggested using the average predicted properties. 
% %
% However, the accuracy of IP property predicted by one trained model is good enough to optimize the antioxidants.
% %
% In the IP predictor, only one 3D conformer is generated for each training sample.

\section{Experiments \& Results}
\subsection{Antioxidant Optimization Experiment}\label{sec:anti_exp} 
% The models are trained with a proprietary antioxidant data set \cite{anti_dataset}, which contains more than 500 antioxidant molecules. 
% 256 molecules are randomly selected from the data set to train the models in Table~\ref{table:summary_models}.
% The models in Table~\ref{table:summary_models} are trained with a proprietary antioxidant data set \cite{anti_dataset}, which contains more than 500 antioxidant molecules. 256 randomly selected antioxidants are used to train the models.
% The models in Table 1 are trained on 256 antioxidants randomly selected from a proprietary antioxidant data set \cite{anti_dataset},  
% which contains more than 500 antioxidant molecules.
The models in Table~\ref{table:summary_models} are trained on a random subset of 256 antioxidants that are from a proprietary data set \cite{anti_dataset} of over 500 antioxidant molecules.
\begin{table}[ht]
\begin{tabular}{|c|c|c|c|c|c|c|c|}
\hline
Model      & Initial Mols/Model & Trained Models & Episodes & Nodes & Modification Batch  \\ \hline
Individual & 1            & 256            & 8000     & 1     & 1                \\ \hline
Parallel   & 8            & 32             & 8000     & 1     & 1                \\ \hline
General    & 256          & 1              & 250      & 4     & 4                \\ \hline
Fine-Tuned & 1            & 256            & 200      & 1     & 1                 \\ \hline 
\end{tabular}

\caption{This table summarizes the number of initial molecules, computation resources, and episodes to train the models. 
%
% Each individual model is trained with 1 initial molecule, and the model is also used to optimize the initial molecule.
%
% There are 256 trained individual models to optimize the selected antioxidants.
%
% Similarly, the parallel model is trained with 8 initial molecules on one node.
%
% The general model is trained on 4 nodes and 
Each node has 4 $\times$ Tesla A100 GPUs.
%
% It will optimize 256 initial antioxidants simultaneously and fewer episodes are needed. % to converge.
%
The fine-tuning process starts with the pre-trained general model and needs 200 extra episodes.
More parameters are in Appendix~\ref{app:expr_para}.
}
\label{table:summary_models}
\end{table}

\paragraph{Summary of Expected Properties}\label{sec:summary_expected_properties}
The optimized molecules are expected to have the following properties or constraints:
% \begin{enumerate}[label=(\Alph*)]
% \item 
(A) 
lower BDE than the original antioxidants ( $ < $ 76 kcal/mol).
% \item 
(B) 
higher IP than the original antioxidants ( $ > $ 145 kcal/mol).
% \item 
(C) 
fewer atoms and bonds.
% \item 
(D) 
similar but not identical to the original antioxidants.
% \item 
(E) 
low SA score.
% \end{enumerate}
A, B, and C are satisfied in the reward function (Equation~\ref{Eq:reward}).% in Section~\ref{sec:normalized_reward_and_reward_function}.
%
% An extra filter script is needed for D and E, which will filter out molecules that are not similar to other antioxidants or have a high SA score.%, the details are in Section~\ref{sec:filter_script}.
The extra filter script in Section~\ref{sec:filter_script} is needed for D and E. %, which will filter out molecules that are not similar to other antioxidants or have a high SA score.

\paragraph{Optimization Failure Rate}
% optimization failure rate.
An optimization is successful if the BDE of the generated molecule is less than 76 kcal/mol and the IP is greater than 145 kcal/mol. Otherwise, the optimization fails and training resources are wasted.
The optimization failure rate (OFR) is defined in Equation~\ref{Eq:ofr} where $S$ is the number of successful optimizations and $A$ is the total number of optimizations.
\begin{equation}
\text{OFR} = 1 - S / A
\label{Eq:ofr}
\end{equation}

% \subsection{Rewards of Different Models}
% \subsection{Rewards of Training}

\subsection{Optimization Results of Different Models}

\begin{figure}[ht]
\centering % <-- added
\begin{subfigure}{}\label{fig:compare_rewards_07_27}
  \includegraphics[trim=0 0 0 0,clip,width=0.47\textwidth]{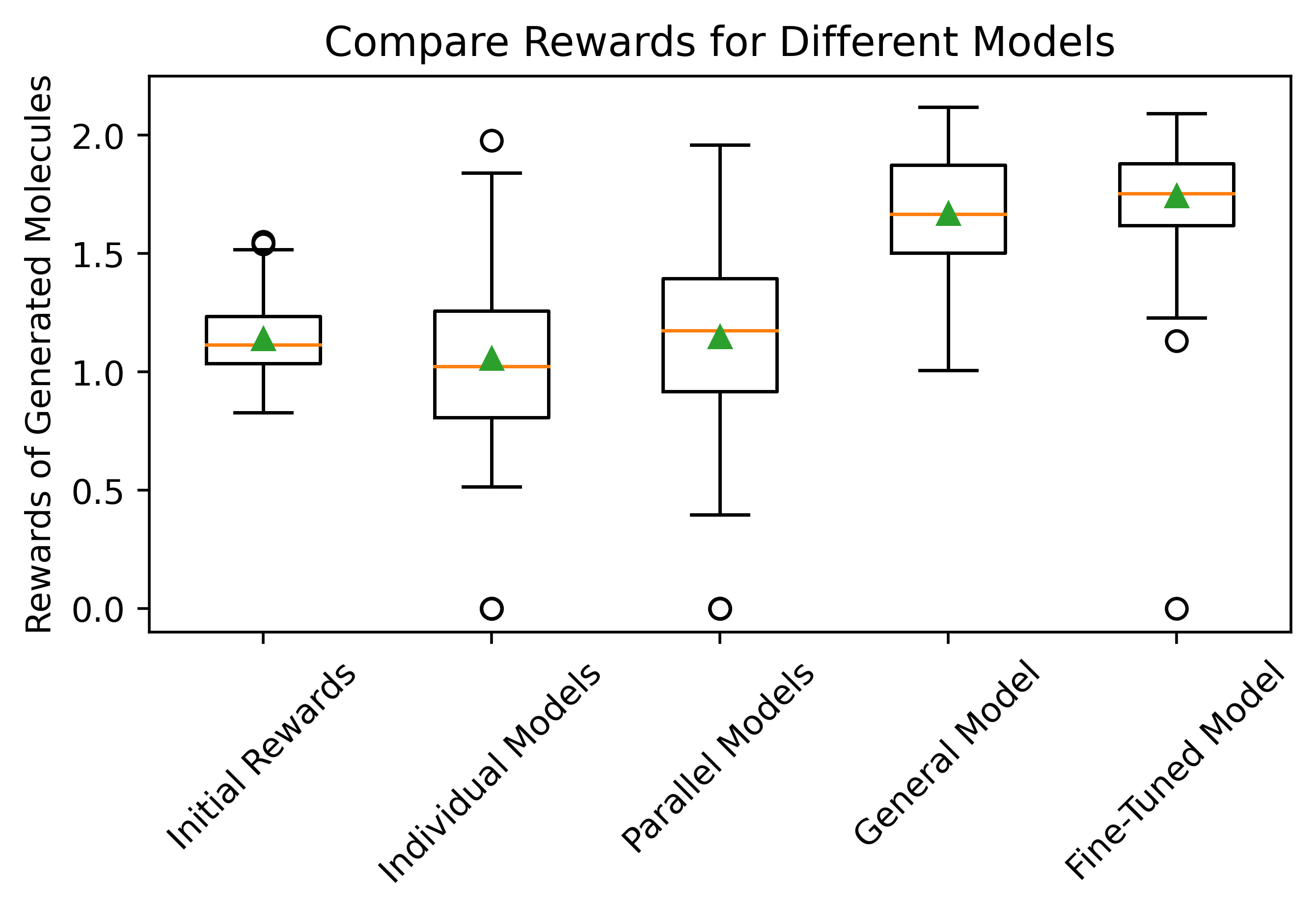}\hfil
\end{subfigure}
\begin{subfigure}{}\label{fig:ofr}
  \includegraphics[trim=0 0 0 0,clip,width=0.47\textwidth]{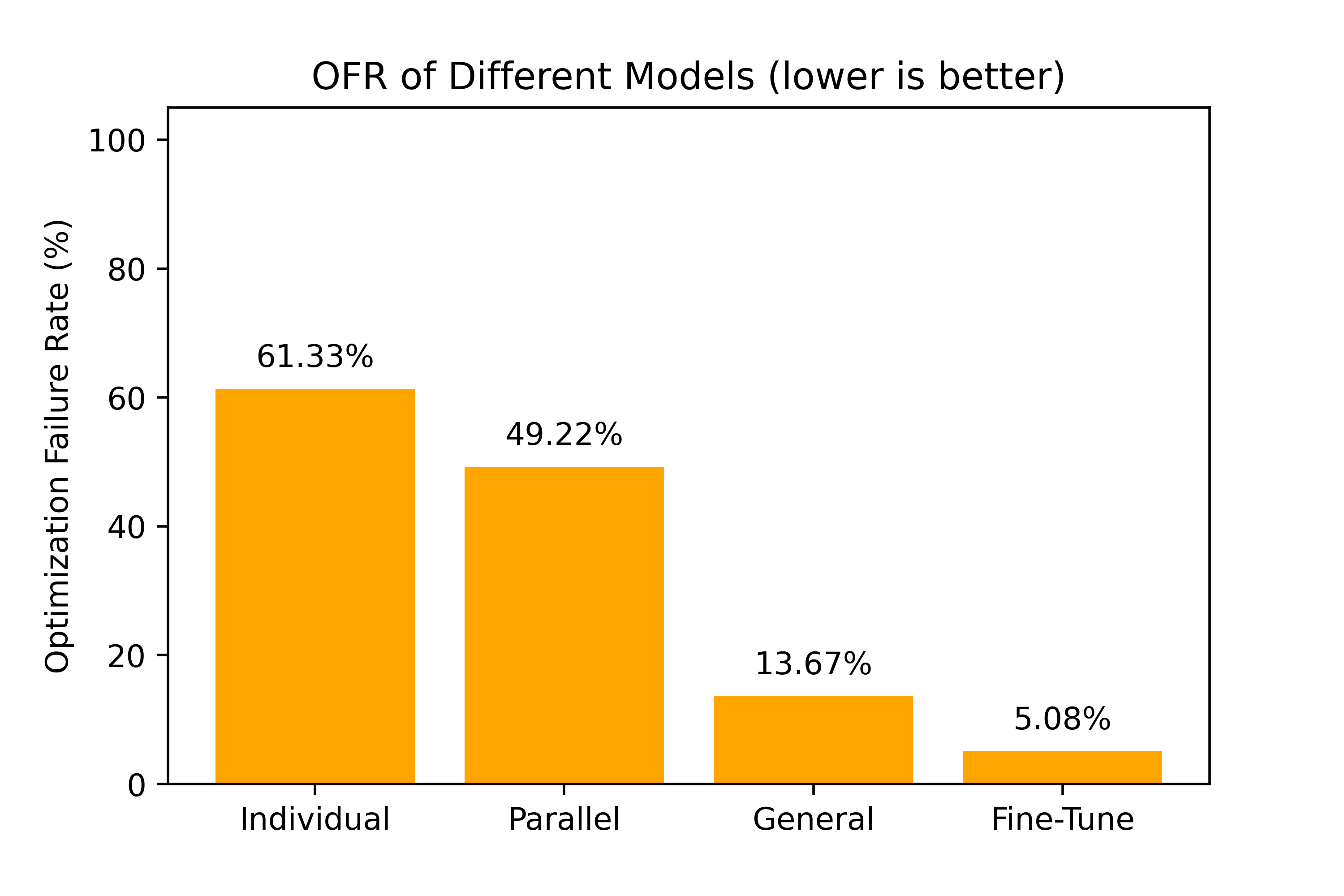}\hfil
\end{subfigure}
\caption{Left: This figure compares the rewards of antioxidants optimized by 4 different models.
    The rewards of a few molecules are out of range (0.0 - 3.0) because of invalid 3D conformers. % or abnormal predicted IP values.
    Their rewards are set to 0. % so their extreme values won't influence the average reward too much. 
    %
    % Note that those unusual rewards can not be ignored because they are still failed optimizations.
    %
    Right: The figure compares the OFR of different models. 
    % The general model and fine-tuned models have much more successful optimization than individual and parallel models.
}
\label{fig:compare_rewards_ofr} 
\end{figure}

\begin{figure}[ht]
\centering % <-- added
\begin{subfigure}{}
  \includegraphics[trim=0 0 0 0,clip,width=0.47\textwidth]{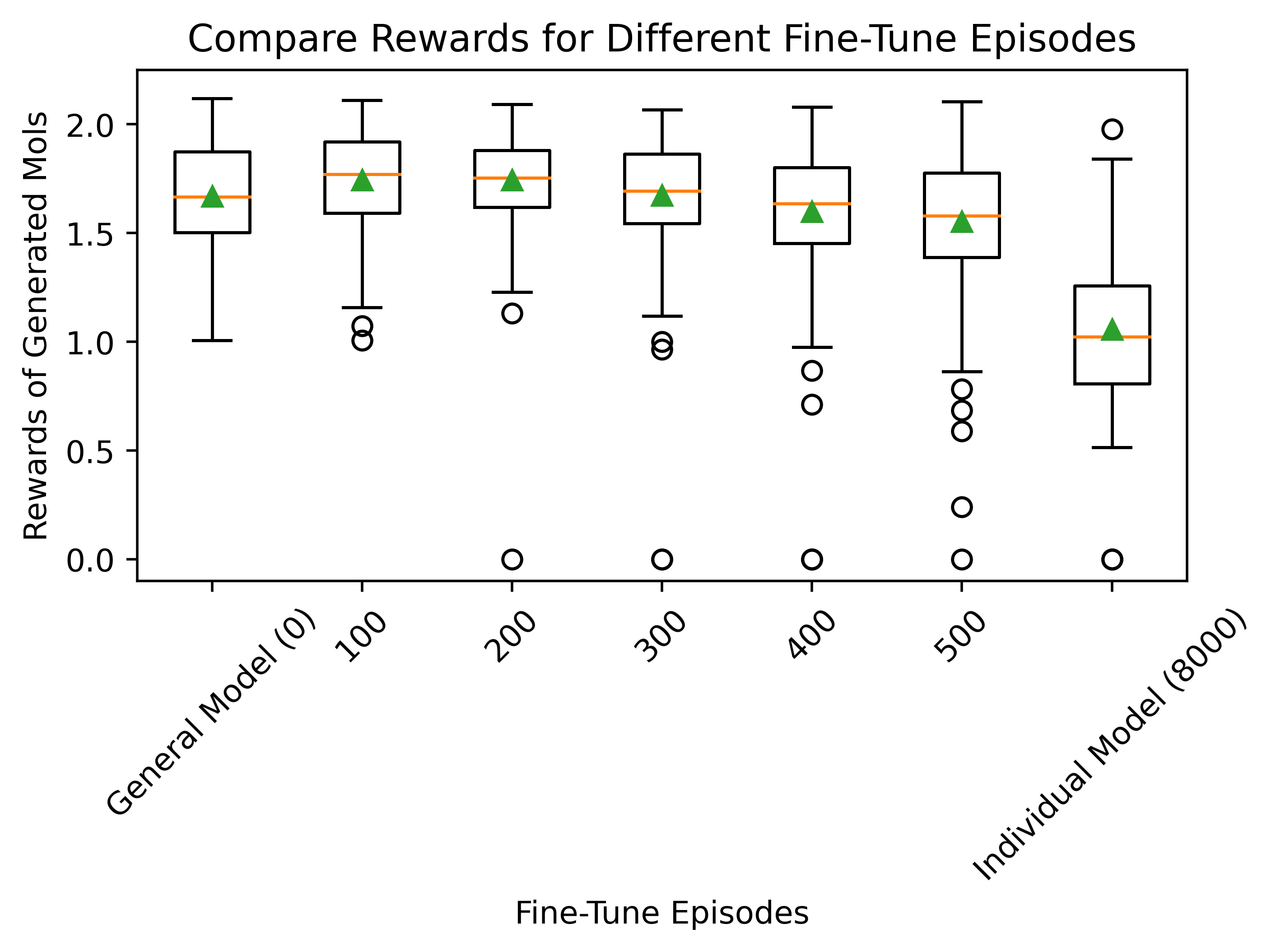}\hfil
\end{subfigure}
\begin{subfigure}{}
  \includegraphics[trim=0 0 0 0,clip,width=0.47\textwidth]{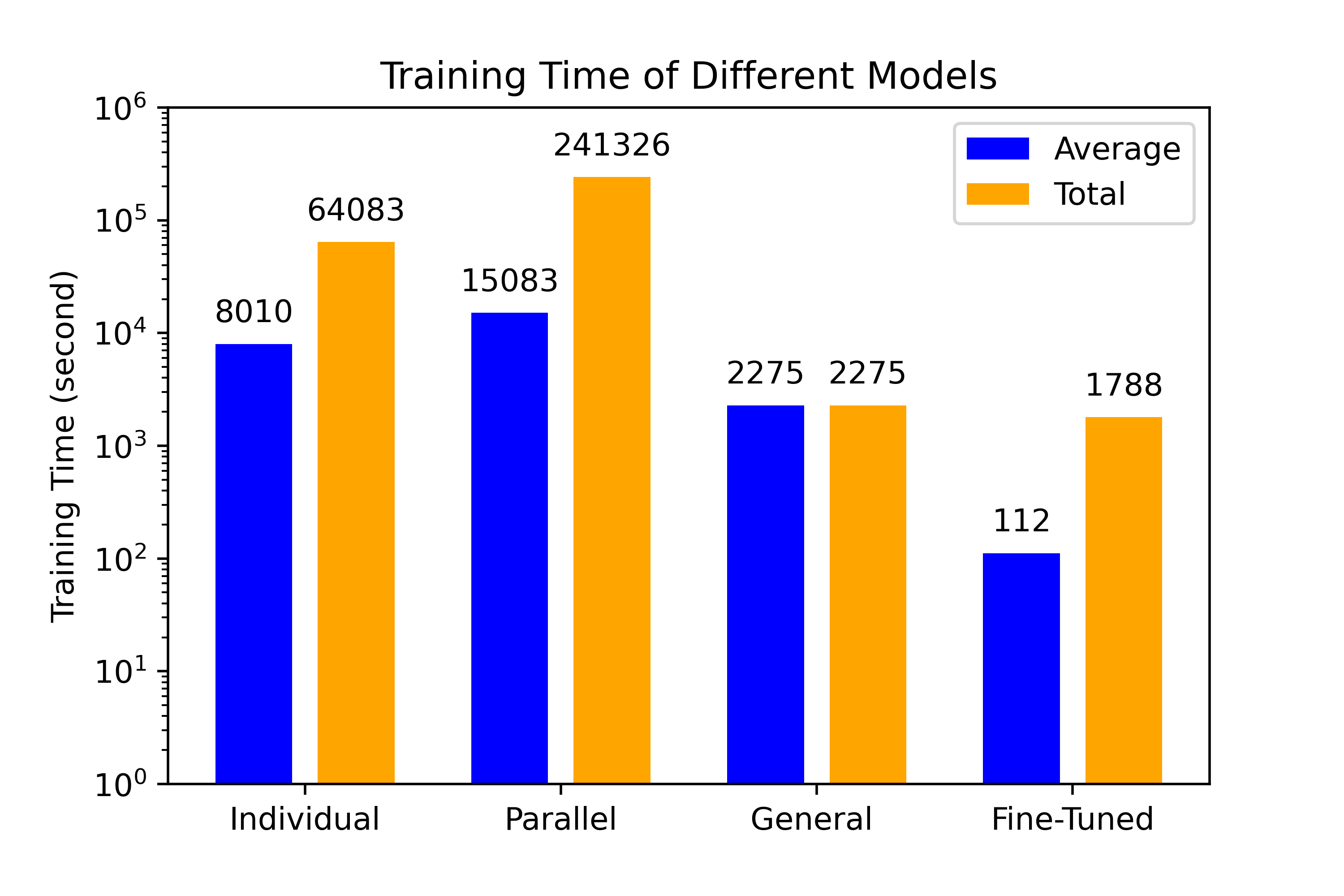}\hfil
\end{subfigure}
\caption{The left figure compares rewards of different fine-tune episodes. 
    %
    % The fine-tuned models are able to optimize the irregular antioxidants within 100 or 200 extra episodes.
    %
    % The lower bounds of fine-tuned models are higher than the general model, resulting in lower OFR.
    %
    % When the models are trained with more and more fine-tuned episodes, the fine-tuned models are eventually individual models.
    %
    The right figure shows the computation time for training each model. The episodes of each model are in Table~\ref{table:summary_models}.
    The blue bars are the average training time per model and the orange bars are the total computation time to train all models while using all 4 nodes.
}
\label{fig:compare_fine_tune_comp_time} 
\end{figure}

Figure~\ref{fig:compare_rewards_ofr} shows the rewards of antioxidants optimized by the individual, parallel, general, and fine-tuned models.
In the left figure, the rewards of individuals and parallel models are not improved compared to initial molecules.
61.33\% and 49.22\% of their optimizations are failed.
The rewards of the general model are significantly higher than the rewards of initial molecules, individual models, and parallel models,
and the OFR is also much lower than the individual models and parallel models.
For the fine-tuned models, the rewards are only slightly improved, % that is the lower bound of fine-tuned models is higher than that of the general model. 
% The OFR of fine-tuned models is further reduced.
and the OFR of fine-tuned models is further reduced.
The results indicate that most antioxidants are successfully optimized by the general model and fine-tuned models.
In Figure~\ref{fig:compare_fine_tune_comp_time}, when the models are trained with more and more fine-tuned episodes, the fine-tuned models are eventually individual models.
The results show that 100 or 200 extra episodes are enough for the fine-tuned models to optimize the irregular antioxidants.

% The general model is trained with 256x and 32x more initial antioxidants than individual and parallel models.
% This helps the general model to naturally explore a broader molecular space without being limited to the vicinity of individual molecules, and it is less prone to getting stuck in local optima. 
% Another advantage of the general model lies in its ability to achieve much lower OFR than individual and parallel models.

% \paragraph{Results of Fine-Tuned Models}
% % The results of fine-tuned models look similar to the general model.
% % %
% In Figure~\ref{fig:compare_rewards_ofr}, the lower bound of fine-tuned models is higher than that of the general model,
% %
% and the fine-tuned models also have lower OFR.
% %
% The fine-tuned models are actually specialized for different molecules, which allows them to better optimize outlier molecules, leading to higher optimization success rates.
% %
% However, as shown in Figure~\ref{fig:compare_fine_tune_comp_time}, if too many (> 300) fine-tune episodes are run, the model tends to be individual models and loses its general advantages.
% %

\paragraph{Computation Time of Training and Fine-tuning}
In Figure~\ref{fig:compare_fine_tune_comp_time}, the computation time of the general model is 3.5x and 6.6x faster than individual models and parallel models, while the general model is more powerful in optimizing antioxidants.
Note that it needs only 1 general model to optimize the 256 training molecules, but it needs 256 individual models and 32 parallel models to optimize all 256 molecules. 
% The training time in Figure~\ref{fig:compare_fine_tune_comp_time} is the average time for training 1 model.
% It means that the total training and optimization time of individual and parallel models will be 8x and 16x slower than the average training time on 4 nodes.
% The individual and parallel models are 28.1x and 106x slower than the general model.
% The above performance improvement does not include general performance optimizations, because they use different reward functions.
Although the 4 nodes with 16 A100 GPUs could train 16 individual models or 4 parallel models simultaneously, 
their optimizations of 256 molecules are still 28.1x and 106x slower than the general model. %, to complete the optimization of 256 molecules, 
Figure~\ref{fig:compare_fine_tune_comp_time} also shows that the extra computation time of fine-tuning is trivial compared to training an initial model from scratch.

% \subsection{Optimization of Unseen and Public Molecules}
% \paragraph{Optimization of Unseen Proprietary Antioxidants}
\subsection{Optimization of Unseen Proprietary Antioxidants}
\begin{figure}[ht]
\centering % <-- added
\begin{subfigure}{}\label{fig:compare_rewards_07_27_test}
  \includegraphics[trim=0 0 0 0,clip,width=0.47\textwidth]{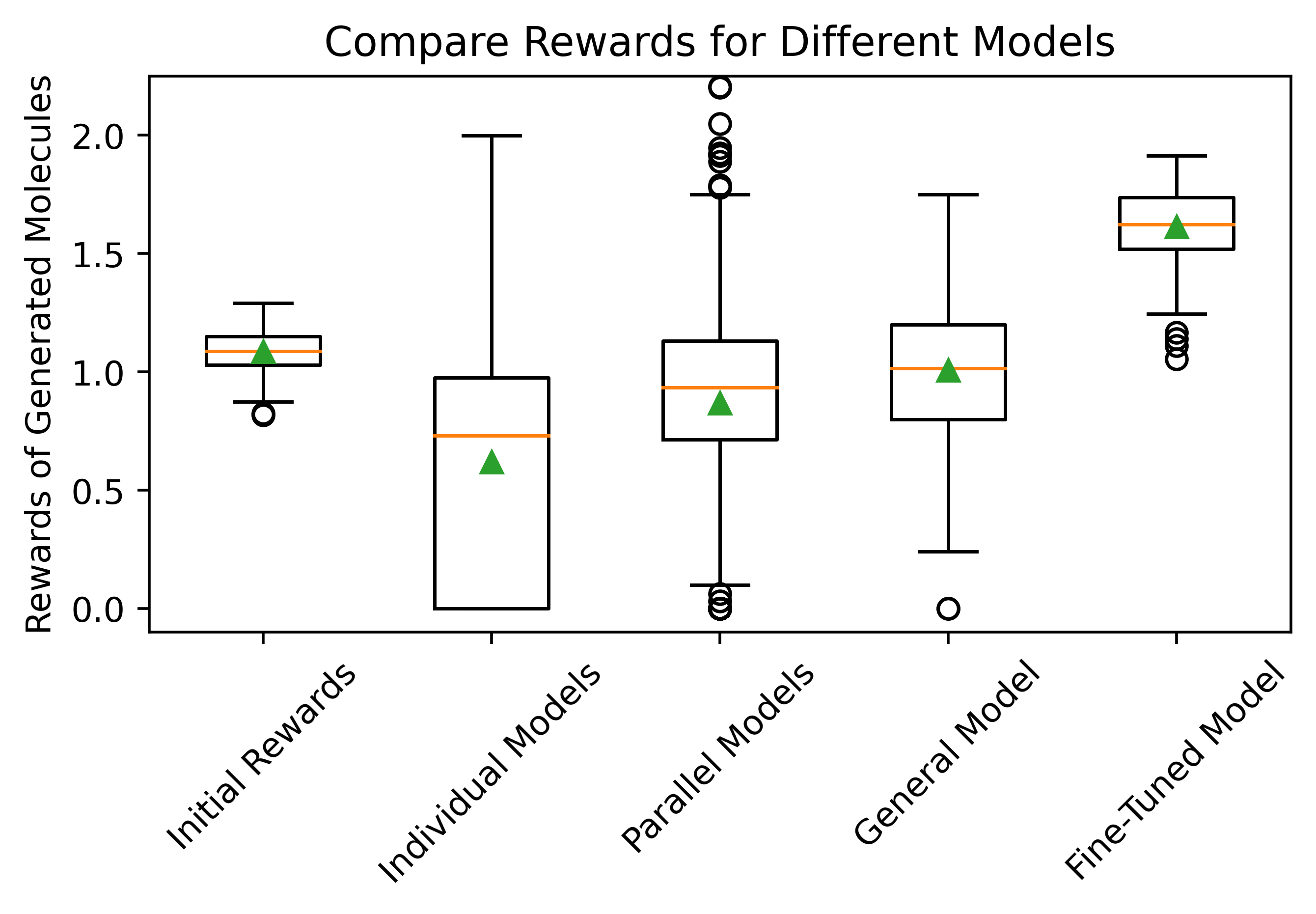}\hfil
\end{subfigure}
\begin{subfigure}{}\label{fig:ofr_test}
  \includegraphics[trim=0 0 0 0,clip,width=0.47\textwidth]{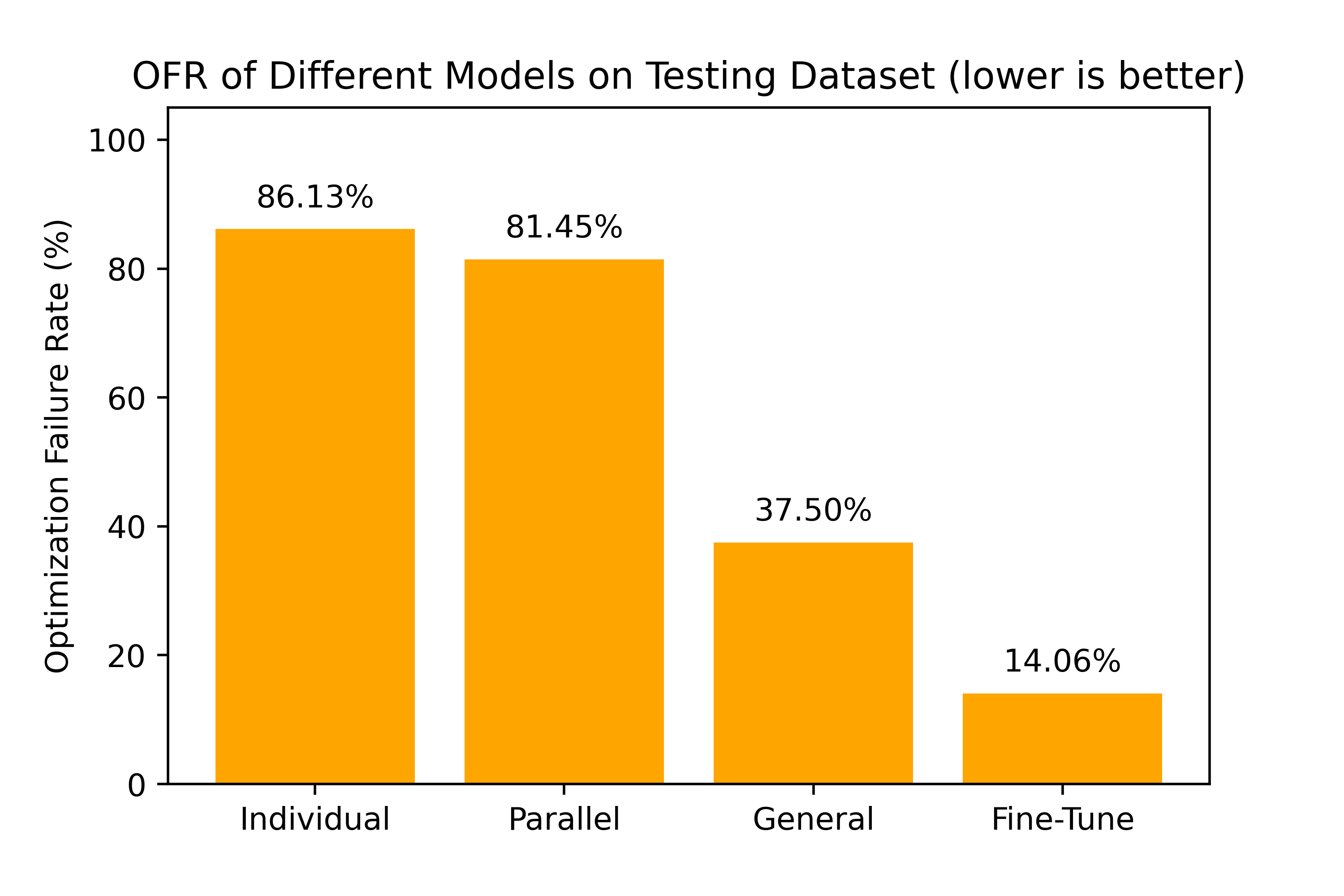}\hfil
\end{subfigure}
\caption{Left: This figure compares the rewards of 128 test antioxidants optimized by 4 different models.
    Right: The figure compares the OFR of the test antioxidants. 
    % The general model is only able to optimize 62.50\% test antioxidants, 
    % The individual models and parallel models fail to optimize most unseen antioxidants,
    % while the fine-tuned models are still able to optimize most antioxidants.
    % The general model and fine-tuned models still outperform individual and parallel models.
}
\label{fig:compare_rewards_ofr_test} 
\end{figure}

% The proprietary antioxidant data set has 500 molecules, and 256 of them are used to train the models.
128 testing molecules are randomly selected from the rest antioxidants and are optimized by the trained models.
16 independent models and 16 parallel models are also randomly picked from the 256 and 32 trained models.
These models, the general model, and the fine-tuning models are used to optimize all 128 testing antioxidants. 
Their optimized rewards and OFR are shown in Figure~\ref{fig:compare_rewards_ofr_test}.
Individual models and parallel models are not able to optimize unseen molecules. 
The general model still outperforms individual and parallel models,
but the effect of fine-tuning is more significant for optimizing unseen molecules.
The results show that the proposed DA-MolDQN model could still well optimize the unseen antioxidants,
under the help of 200 fine-tuning episodes.

% \subsection{Validation of ML Results with DFT Simulations}
\subsection{Proposed Antioxidants}
\begin{figure}[ht]
\centering % <-- added
\begin{subfigure}{}\label{fig:compare_bde_ip}
  \includegraphics[trim=0 0 0 0,clip,width=0.47\textwidth]{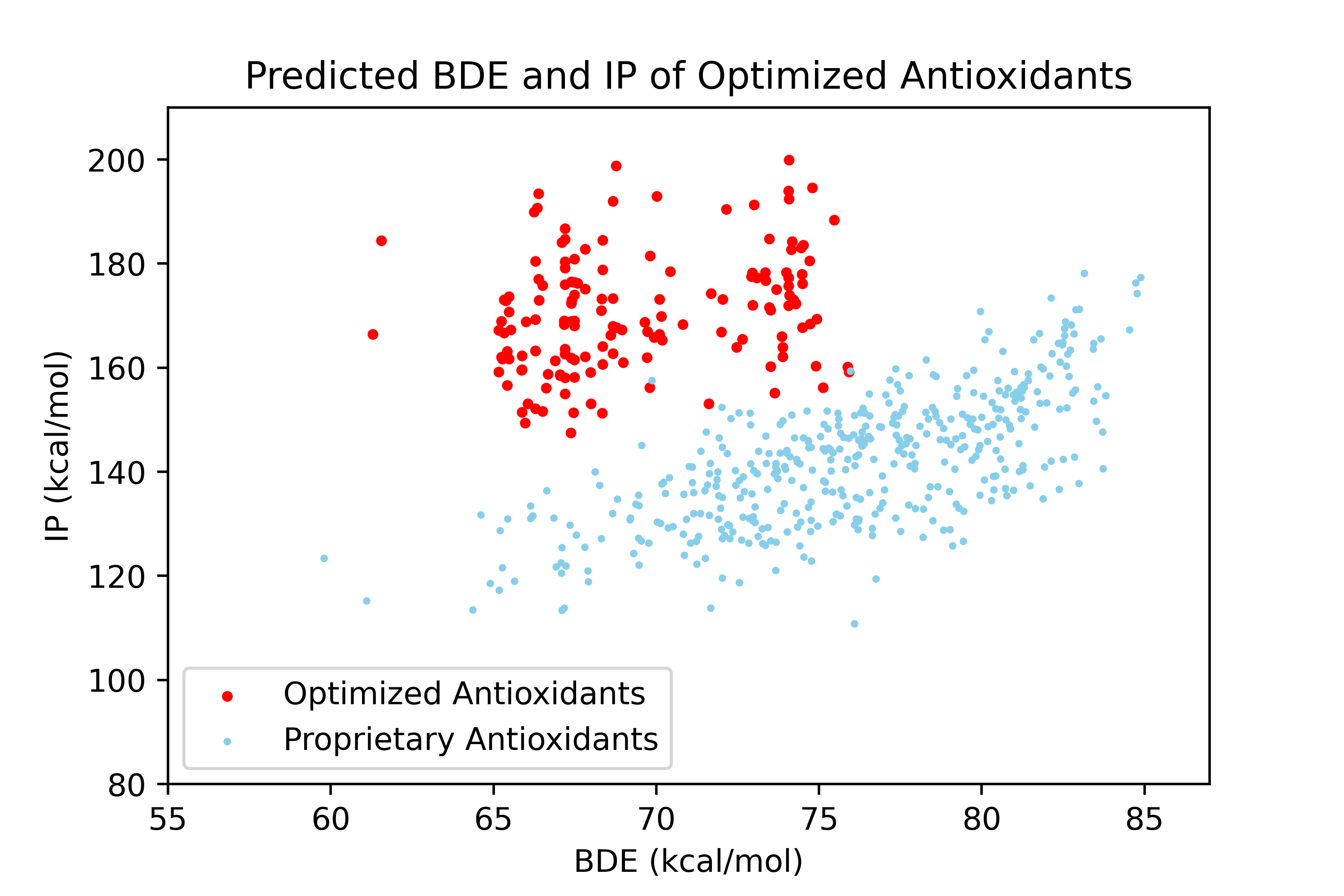}\hfil
\end{subfigure}
\begin{subfigure}{}\label{fig:compare_sim_sas}
  \includegraphics[trim=0 0 0 0,clip,width=0.47\textwidth]{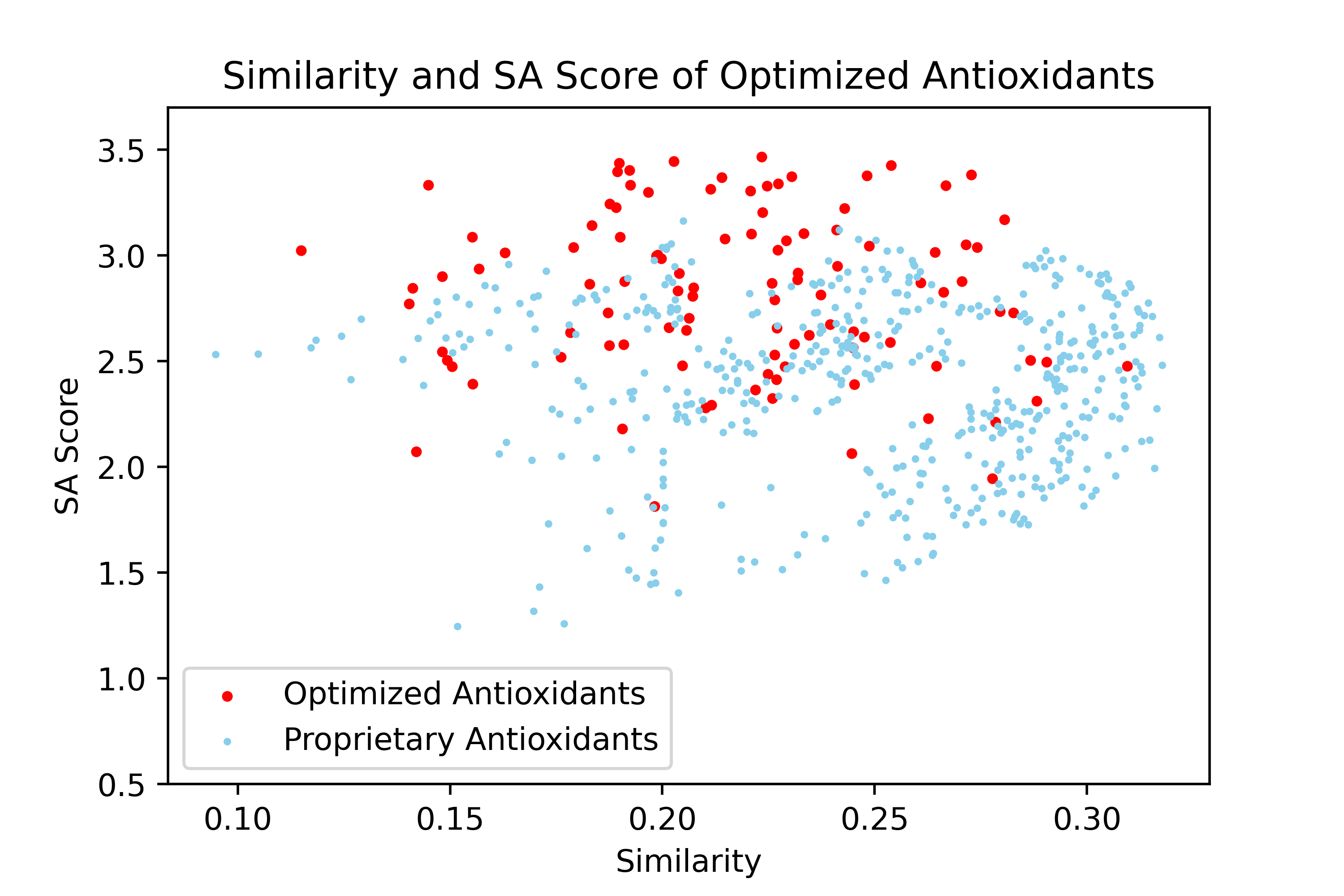}\hfil
\end{subfigure}
\caption{
Left: The predicted BDE and IP of optimized molecules (red) are better than the initial molecules (blue).   
Right: The Similarity and SA Score of generated molecules (red) are close to the initial molecules (blue).
}
\label{fig:compare_bs} 
\end{figure}
\paragraph{Optimized BDE and IP Properties}
Figure~\ref{fig:compare_bs} compares the initial molecules and the molecules generated by the trained  DA-MolDQN model. 
The molecules are filtered out if the properties required in Section~\ref{sec:summary_expected_properties} are not satisfied.
The figure shows that the generated molecules have significantly lower BDE and higher IP, compared to all initial molecules in the antioxidant data set. 
The proposed molecules have both good BDE and IP properties, indicating that the optimizer successfully balances and optimizes these chemical properties. 
\paragraph{Similarity and SA Score}
Because similarity and SA score are not included in the RL reward function, there are some optimized molecules that have SA scores greater than 3.5. 
These optimizations are still counted as successful optimizations, even though they are unlikely to be good antioxidant candidates in practice.
The distribution of the remaining optimized molecules in Figure~\ref{fig:compare_bs} is close to the original antioxidants, which is exactly what the chemist expected.

\paragraph{Example of Proposed Antioxidant and Generating Path}
% The molecules in the proprietary antioxidant data set and the molecules generated from them are commercially confidential. 
% Instead of directly showing them, public data is used as a proxy. 
% Some initial molecules with similar BDE and IP are searched in the ChEMBL data set \cite{chembl}, and then the experiments are replayed for the public molecules.
% The experiment results are similar to proprietary molecules and the proposed antioxidants are shown in Figure~\ref{fig:gm_public}. More molecules are shown in Appendix~\ref{app:public_gm_all}.
The molecules in the proprietary antioxidant data set and the molecules generated from them are commercially confidential. 
The experiments are replayed for some public molecules in the ChEMBL \cite{chembl} and AODB \cite{aodb} data sets.
An example of the optimization path is shown in Figure~\ref{fig:gm_public}. More molecules are shown in Appendix~\ref{app:public_gm_all}.
\begin{figure}[ht] % add ht
    \centering
    \begin{subfigure}{}
      \includegraphics[trim=0 0 0 0,clip,width=0.15\textwidth]{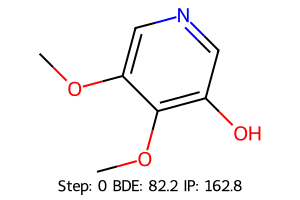}\hfil
    \end{subfigure}
    \begin{subfigure}{}
      \includegraphics[trim=0 0 0 0,clip,width=0.15\textwidth]{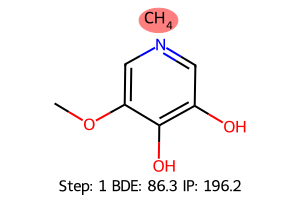}\hfil
    \end{subfigure}
    \begin{subfigure}{}
      \includegraphics[trim=0 0 0 0,clip,width=0.15\textwidth]{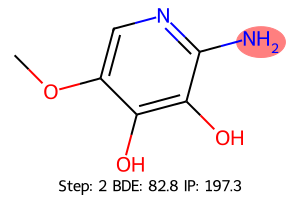}\hfil
    \end{subfigure}
    \begin{subfigure}{}
      \includegraphics[trim=0 0 0 0,clip,width=0.15\textwidth]{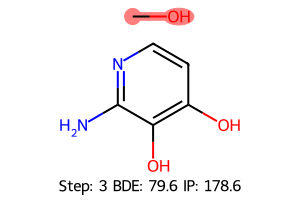}\hfil
    \end{subfigure}
    \begin{subfigure}{}
      \includegraphics[trim=0 0 0 0,clip,width=0.15\textwidth]{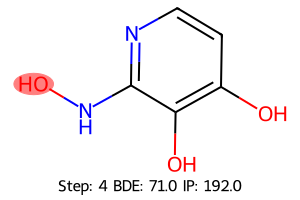}\hfil
    \end{subfigure}
    \begin{subfigure}{}
      \includegraphics[trim=0 0 0 0,clip,width=0.15\textwidth]{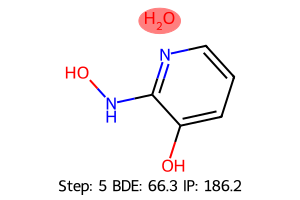}\hfil
    \end{subfigure}
 
    \caption{The figure shows one optimized antioxidant and its generating path.
    The molecules in Step 0 are the initial molecules and the optimized molecules are in Step 5.
    The modifications are highlighted and the unconnected atoms are removed from the current molecules. 
    }
    \label{fig:gm_public}
\end{figure}

\section{Validation of ML Results with DFT Simulations}
\subsection{DFT validation}
% Geometry of each antioxidant molecules, both molecules of the initial dataset and generated molecules, their radical structures and their radical cation structures were calculated by means of Density Functional Theory (DFT) (\cite{parr1979local}) implemented in the Gaussian 16 computer program using B3LYP hybrid functional with \textit{6-311++G(d,p)} basis set in Polarizable Continuum Model of solvation (\cite{dehkordi2022comparative}) where the solvent, Toluene in the present work, is described by its dielectric constant.
The geometry of both initial molecules and optimized molecules, their radical structures, and their radical cation structures were calculated by means of DFT, which is implemented in the Gaussian 16 \cite{g16} computer program using B3LYP hybrid functional with \textit{6-311++G(d,p)} basis set in Polarizable Continuum Model of solvation \cite{dehkordi2022comparative} where the solvent, Toluene in the present work, is described by its dielectric constant.

The following equations are used to calculate BDE and IP on optimized structure \cite{vakarelska2016monohydroxy,mohajeri2009theoretical}:

\begin{equation}
BDE = H(R^\bullet) + H(H^\bullet) – H(R-H)
\end{equation}
\begin{equation}
IP = H(RH^{+\bullet}) + H (e-) -H (R-H)   
\end{equation}

Where $H(R^\bullet)$ is the enthalpy of the radical formed after abstraction of the hydrogen from the phenol group and after geometry optimization of the structure. $H (H^\bullet)$ is the enthalpy of a single H atom. $H(R-H)$ is the enthalpy of the neutral molecule. $H(RH^{+\bullet})$ is the enthalpy of the radical cation formed when the most excited electron is abstracted and $H(e-)$ is the enthalpy of an electron. Enthalpy of the hydrogen atom is $-312,44$ kcal/mol \cite{mohajeri2009theoretical} and enthalpy of the electron is $-55,61$ kcal/mol \cite{vakarelska2016monohydroxy}.

7 proposed molecules are selected from the antioxidants optimized by the parallel models for evaluation with DFT simulations. The DFT results show that the proposed molecules have significantly improved properties and are close to the predicted values.
%
%
%DFT simulation result of the general model is underway.
%The molecules were simulated with DFT to calculate the energy required for antioxidant ionization. 
%The ionization split the molecule into radicals. The energy state of resulting molecules and radicals was evaluated.
The table below shows the comparison of predicted and DFT results, for BDE and IP.
The results indicate that the error is within tolerance for optimizing these molecules.
The details of DFT results are in Appendix~\ref{app:dft_validation}

Figure~\ref{fig:dft_compare} shows the classification of the proposed 7 molecules to their stability and performance both with DFT and DA-MolDQN algorithms.
The comparison shows that 5 of out 7 molecules match the classification for stability and performance.

\begin{figure}[ht] %add ht
    \centering
    \includegraphics[trim=0 0 0 0,clip,width=0.9\textwidth]{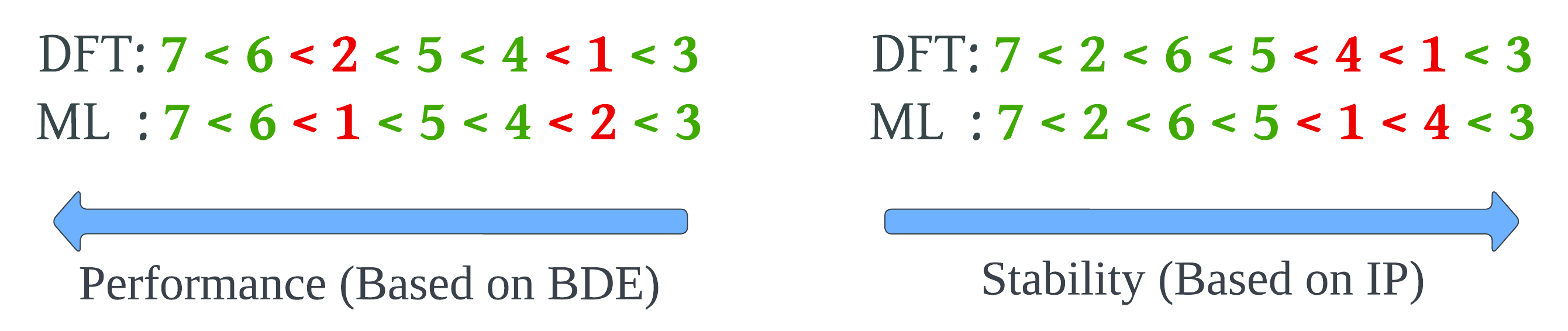}
    \caption{The figure shows the optimized antioxidants ordered by their properties.
    The DFT and ML properties are generated by DFT simulation and the machine learning predictors.  
    }
    \label{fig:dft_compare}
\end{figure}

\section{Discussion}\label{sec:discussion}
% Our model achieved significantly better optimization ability and required much less training time. This is mainly due to the distributed nature of the model, which is not limited by the resources of a single computing node and incorporates batched modification. These features allow the general model to explore a wider range of molecular space freely. The distributed training framework can not only be used to optimize antioxidants efficiently but also to optimize various drug-like molecules by introducing other property predictors. 
% However, there are still some limitations in this study. Firstly, our study relies on RL to learn to avoid invalid 3D conformers, which means that antioxidant optimization is not performed in the early stages of training, leading to reduced learning efficiency. Secondly, there is a lack of comparison with other molecular optimization frameworks in the experiments. Further research, after obtaining the latest DFT validation results, may help to address these limitations.
The distributed training algorithm and the batched modification trained the model with many more initial molecules, compared to the MolDQN and MT-MolDQN.
The general model explores a wider range of the molecular universe, 
resulting in the RL agent learning from various candidate antioxidants.
The experiment results of both training and testing data sets show that the DA-MolDQN achieved significantly better optimization ability than the previous work. %MolDQN and MT-MolDQN.
The effectiveness of DA-MolDQN was proved by the DFT validation results and the fact that we did find some of the discovered antioxidants in PubChem and AODB datasets.
The general models with 512 and 1024 initial molecules are also tested, there is no significant improvement compared to the model with 256 molecules.

% At the beginning, the estimated computation time for optimizing an antioxidant will be over 16 days, 
% which are obviously too slow to be commercially and practically applicable.
By speeding up the modifications, fingerprinting, and property predictions, 
the training performance of DA-MolDQN is greatly improved and the distributed script could optimize 256 antioxidants within an hour, 
instead of taking 16 days to optimize 1 molecule.
% and required much less training time.
The distributed training framework of DA-MolDQN can be used to optimize various drug-like molecules by introducing other property predictors.
The performance optimization methods could also be helpful to other applications.

\section{Conclusions \& Future Work}
In this paper, state-of-the-art BDE and IP predictors were integrated and the training performance was greatly improved. %into the distributed antioxidant optimizer.
The molecules proposed by DA-MolDQN have similarity and synthesizability scores close to the original molecules 
while their BDE and IP properties are improved,
and the results are validated by DFT and public datasets.
In the future, more optimization will be applied so that the general model for 256-1024 molecules can be trained using less computational resources.
Another approach is to find a method which can avoid 100\% invalid 3D conformers without unnecessary RL actions. Additional DFT validations are underway.

% In this paper, state-of-the-art BDE and IP predictors were integrated into the distributed antioxidant optimizer.
% The molecules proposed by DA-MolDQN have similarity and synthesizability scores close to the original molecules while improving BDE and IP properties.
% The training time of DA-MolDQN was also significantly improved.
% Currently, DFT validation of the newest molecules proposed by the general model is underway.

% to do: update dft validation
% to do : add found gms in our dataset
% to do: add increased initial mol
% to do: scale problem was solved compared to the workshop paper. mention the python-> c++, fp
% caching, compared to the workshop paper.
% invalid 3d conformers
% add 1024 mols if need

\bibliography{ai4mat_workshop}

\begin{thebibliography}{40}
\providecommand{\natexlab}[1]{#1}
\providecommand{\url}[1]{\texttt{#1}}
\expandafter\ifx\csname urlstyle\endcsname\relax
  \providecommand{\doi}[1]{doi: #1}\else
  \providecommand{\doi}{doi: \begingroup \urlstyle{rm}\Url}\fi

\bibitem[Alimova et~al.(2021)Alimova, Kholikova, Kholova, and
  Karimova]{anti_application_2}
Z.~X. Alimova, N.~A. Kholikova, S.~O. Kholova, and K.~G. Karimova.
\newblock Influence of the antioxidant properties of lubricants on the wear of
  agricultural machinery parts.
\newblock \emph{IOP Conference Series: Earth and Environmental Science},
  868\penalty0 (1):\penalty0 012037, oct 2021.
\newblock \doi{10.1088/1755-1315/868/1/012037}.
\newblock URL \url{https://dx.doi.org/10.1088/1755-1315/868/1/012037}.

\bibitem[Amran et~al.(2022)Amran, Bello, and Ruslan]{bde_r2}
N.~A. Amran, U.~Bello, and M.~S.~H. Ruslan.
\newblock The role of antioxidants in improving biodiesel’s oxidative
  stability, poor cold flow properties, and the effects of the duo on engine
  performance: A review.
\newblock \emph{Heliyon}, page e09846, 2022.

\bibitem[Bajusz et~al.(2015)Bajusz, R{\'a}cz, and H{\'e}berger]{tanimoto}
D.~Bajusz, A.~R{\'a}cz, and K.~H{\'e}berger.
\newblock Why is tanimoto index an appropriate choice for fingerprint-based
  similarity calculations?
\newblock \emph{Journal of cheminformatics}, 7\penalty0 (1):\penalty0 1--13,
  2015.

\bibitem[Blaney and Dixon(1994)]{embedMolecule1}
J.~M. Blaney and J.~S. Dixon.
\newblock \emph{Distance Geometry in Molecular Modeling}, pages 299--335.
\newblock John Wiley \& Sons, Ltd, 1994.
\newblock ISBN 9780470125823.
\newblock \doi{https://doi.org/10.1002/9780470125823.ch6}.
\newblock URL
  \url{https://onlinelibrary.wiley.com/doi/abs/10.1002/9780470125823.ch6}.

\bibitem[Born et~al.(2021)Born, Manica, Oskooei, Cadow, Markert, and
  {Rodríguez Martínez}]{rl_mol_2}
J.~Born, M.~Manica, A.~Oskooei, J.~Cadow, G.~Markert, and M.~{Rodríguez
  Martínez}.
\newblock Paccmannrl: De novo generation of hit-like anticancer molecules from
  transcriptomic data via reinforcement learning.
\newblock \emph{iScience}, 24\penalty0 (4):\penalty0 102269, 2021.
\newblock ISSN 2589-0042.
\newblock \doi{https://doi.org/10.1016/j.isci.2021.102269}.
\newblock URL
  \url{https://www.sciencedirect.com/science/article/pii/S2589004221002376}.

\bibitem[ChEMBL(2022)]{chembl}
ChEMBL.
\newblock "\url{https://www.ebi.ac.uk/chembl/}", 2022.

\bibitem[{da Silva} et~al.(2006){da Silva}, Chen, and Bozzelli]{bde_oh_88_90}
G.~{da Silva}, C.-C. Chen, and J.~W. Bozzelli.
\newblock Bond dissociation energy of the phenol oh bond from ab initio
  calculations.
\newblock \emph{Chemical Physics Letters}, 424\penalty0 (1):\penalty0 42--45,
  2006.
\newblock ISSN 0009-2614.
\newblock \doi{https://doi.org/10.1016/j.cplett.2006.04.022}.
\newblock URL
  \url{https://www.sciencedirect.com/science/article/pii/S0009261406004982}.

\bibitem[Daguet-Schott()]{bde_ms}
C.~Daguet-Schott.
\newblock New methods to characterize fuels’ oxidation and screening of novel
  antioxidants.
\newblock Master's thesis, Department of Chemistry Technical University of
  Denmark.

\bibitem[DAI et~al.(2019)DAI, Akhiyarov, Alami, Pantano, Araya-Polo, and
  Pereira]{mt-moldqn}
Z.~DAI, D.~Akhiyarov, R.~Alami, D.~Pantano, M.~Araya-Polo, and C.~Pereira.
\newblock Multi-task deep reinforcement learning for molecular optimization.
\newblock \emph{ELLIS Machine Learning for Molecule Discovery Workshop}, 2019.

\bibitem[Dehkordi et~al.(2022)Dehkordi, Asgarshamsi, Fassihi, and
  Zborowski]{dehkordi2022comparative}
M.~M. Dehkordi, M.~H. Asgarshamsi, A.~Fassihi, and K.~K. Zborowski.
\newblock A comparative dft study on the antioxidant activity of some novel
  3-hydroxypyridine-4-one derivatives.
\newblock \emph{Chemistry \& Biodiversity}, 19\penalty0 (3):\penalty0
  e202100703, 2022.

\bibitem[Deng et~al.(2023)Deng, Chen, Sun, and Wang]{aodb}
W.~Deng, Y.~Chen, X.~Sun, and L.~Wang.
\newblock Aodb: A comprehensive database for antioxidants including small
  molecules, peptides and proteins.
\newblock \emph{Food Chemistry}, 418:\penalty0 135992, 2023.
\newblock ISSN 0308-8146.
\newblock \doi{https://doi.org/10.1016/j.foodchem.2023.135992}.
\newblock URL
  \url{https://www.sciencedirect.com/science/article/pii/S030881462300609X}.

\bibitem[Fickinger et~al.(2021)Fickinger, Hu, Amos, Russell, and
  Brown]{fine_tune_2}
A.~Fickinger, H.~Hu, B.~Amos, S.~Russell, and N.~Brown.
\newblock Scalable online planning via reinforcement learning fine-tuning.
\newblock \emph{Advances in Neural Information Processing Systems},
  34:\penalty0 16951--16963, 2021.

\bibitem[Frisch et~al.(2016)Frisch, Trucks, Schlegel, Scuseria, Robb,
  Cheeseman, Scalmani, Barone, Petersson, Nakatsuji, Li, Caricato, Marenich,
  Bloino, Janesko, Gomperts, Mennucci, Hratchian, Ortiz, Izmaylov, Sonnenberg,
  Williams-Young, Ding, Lipparini, Egidi, Goings, Peng, Petrone, Henderson,
  Ranasinghe, Zakrzewski, Gao, Rega, Zheng, Liang, Hada, Ehara, Toyota, Fukuda,
  Hasegawa, Ishida, Nakajima, Honda, Kitao, Nakai, Vreven, Throssell,
  Montgomery, Peralta, Ogliaro, Bearpark, Heyd, Brothers, Kudin, Staroverov,
  Keith, Kobayashi, Normand, Raghavachari, Rendell, Burant, Iyengar, Tomasi,
  Cossi, Millam, Klene, Adamo, Cammi, Ochterski, Martin, Morokuma, Farkas,
  Foresman, and Fox]{g16}
M.~J. Frisch, G.~W. Trucks, H.~B. Schlegel, G.~E. Scuseria, M.~A. Robb, J.~R.
  Cheeseman, G.~Scalmani, V.~Barone, G.~A. Petersson, H.~Nakatsuji, X.~Li,
  M.~Caricato, A.~V. Marenich, J.~Bloino, B.~G. Janesko, R.~Gomperts,
  B.~Mennucci, H.~P. Hratchian, J.~V. Ortiz, A.~F. Izmaylov, J.~L. Sonnenberg,
  D.~Williams-Young, F.~Ding, F.~Lipparini, F.~Egidi, J.~Goings, B.~Peng,
  A.~Petrone, T.~Henderson, D.~Ranasinghe, V.~G. Zakrzewski, J.~Gao, N.~Rega,
  G.~Zheng, W.~Liang, M.~Hada, M.~Ehara, K.~Toyota, R.~Fukuda, J.~Hasegawa,
  M.~Ishida, T.~Nakajima, Y.~Honda, O.~Kitao, H.~Nakai, T.~Vreven,
  K.~Throssell, J.~A. Montgomery, {Jr.}, J.~E. Peralta, F.~Ogliaro, M.~J.
  Bearpark, J.~J. Heyd, E.~N. Brothers, K.~N. Kudin, V.~N. Staroverov, T.~A.
  Keith, R.~Kobayashi, J.~Normand, K.~Raghavachari, A.~P. Rendell, J.~C.
  Burant, S.~S. Iyengar, J.~Tomasi, M.~Cossi, J.~M. Millam, M.~Klene, C.~Adamo,
  R.~Cammi, J.~W. Ochterski, R.~L. Martin, K.~Morokuma, O.~Farkas, J.~B.
  Foresman, and D.~J. Fox.
\newblock Gaussian˜16 {R}evision {C}.01, 2016.
\newblock Gaussian Inc. Wallingford CT.

\bibitem[Jette et~al.(2002)Jette, Dunlap, Garlick, and Grondona]{slurm}
M.~Jette, C.~Dunlap, J.~Garlick, and M.~Grondona.
\newblock Slurm: Simple linux utility for resource management.
\newblock 7 2002.
\newblock URL \url{https://www.osti.gov/biblio/15002962}.

\bibitem[Kasote et~al.(2015)Kasote, Katyare, Hegde, and Bae]{anti_bde_ip}
D.~M. Kasote, S.~S. Katyare, M.~V. Hegde, and H.~Bae.
\newblock Significance of antioxidant potential of plants and its relevance to
  therapeutic applications.
\newblock \emph{International journal of biological sciences}, 11\penalty0
  (8):\penalty0 982, 2015.

\bibitem[Li et~al.(2020)Li, Zhao, Varma, Salpekar, Noordhuis, Li, Paszke,
  Smith, Vaughan, Damania, et~al.]{pytorch_ddp}
S.~Li, Y.~Zhao, R.~Varma, O.~Salpekar, P.~Noordhuis, T.~Li, A.~Paszke,
  J.~Smith, B.~Vaughan, P.~Damania, et~al.
\newblock Pytorch distributed: Experiences on accelerating data parallel
  training.
\newblock \emph{arXiv preprint arXiv:2006.15704}, 2020.

\bibitem[Meldgaard et~al.(2021)Meldgaard, Köhler, Mortensen, Christiansen,
  Noé, and Hammer]{rl_mol_4}
S.~A. Meldgaard, J.~Köhler, H.~L. Mortensen, M.-P.~V. Christiansen, F.~Noé,
  and B.~Hammer.
\newblock Generating stable molecules using imitation and reinforcement
  learning.
\newblock \emph{Machine Learning: Science and Technology}, 3\penalty0
  (1):\penalty0 015008, dec 2021.
\newblock \doi{10.1088/2632-2153/ac3eb4}.
\newblock URL \url{https://dx.doi.org/10.1088/2632-2153/ac3eb4}.

\bibitem[Mohajeri and Asemani(2009{\natexlab{a}})]{anti_bde1}
A.~Mohajeri and S.~S. Asemani.
\newblock Theoretical investigation on antioxidant activity of vitamins and
  phenolic acids for designing a novel antioxidant.
\newblock \emph{Journal of Molecular Structure}, 930\penalty0 (1):\penalty0
  15--20, 2009{\natexlab{a}}.
\newblock ISSN 0022-2860.
\newblock \doi{https://doi.org/10.1016/j.molstruc.2009.04.031}.
\newblock URL
  \url{https://www.sciencedirect.com/science/article/pii/S002228600900252X}.

\bibitem[Mohajeri and Asemani(2009{\natexlab{b}})]{mohajeri2009theoretical}
A.~Mohajeri and S.~S. Asemani.
\newblock Theoretical investigation on antioxidant activity of vitamins and
  phenolic acids for designing a novel antioxidant.
\newblock \emph{Journal of Molecular Structure}, 930\penalty0 (1-3):\penalty0
  15--20, 2009{\natexlab{b}}.

\bibitem[Moussa et~al.(2023)Moussa, Wang, Araya-Polo, Back, and
  Dunjko]{anti_dataset}
C.~Moussa, H.~Wang, M.~Araya-Polo, T.~Back, and V.~Dunjko.
\newblock Application of quantum-inspired generative models to small molecular
  datasets.
\newblock \emph{ArXiv}, abs/2304.10867, 2023.
\newblock URL \url{https://api.semanticscholar.org/CorpusID:258291519}.

\bibitem[O'Neil et~al.(1993)O'Neil, O'Neil, and Weikum]{LRU_cache}
E.~J. O'Neil, P.~E. O'Neil, and G.~Weikum.
\newblock The lru-k page replacement algorithm for database disk buffering.
\newblock \emph{SIGMOD Rec.}, 22\penalty0 (2):\penalty0 297–306, jun 1993.
\newblock ISSN 0163-5808.
\newblock \doi{10.1145/170036.170081}.
\newblock URL \url{https://doi.org/10.1145/170036.170081}.

\bibitem[Parr et~al.(1979)Parr, Gadre, and Bartolotti]{parr1979local}
R.~G. Parr, S.~R. Gadre, and L.~J. Bartolotti.
\newblock Local density functional theory of atoms and molecules.
\newblock \emph{Proceedings of the National Academy of Sciences}, 76\penalty0
  (6):\penalty0 2522--2526, 1979.

\bibitem[Popova et~al.(2018)Popova, Isayev, and Tropsha]{rl_mol_1}
M.~Popova, O.~Isayev, and A.~Tropsha.
\newblock Deep reinforcement learning for de novo drug design.
\newblock \emph{Science Advances}, 4\penalty0 (7):\penalty0 eaap7885, 2018.
\newblock \doi{10.1126/sciadv.aap7885}.
\newblock URL \url{https://www.science.org/doi/abs/10.1126/sciadv.aap7885}.

\bibitem[Pospíšil(1992)]{anti_application_3}
J.~Pospíšil.
\newblock Exploitation of the current knowledge of antioxidant mechanisms for
  efficient polymer stabilization.
\newblock \emph{Polymers for Advanced Technologies}, 3\penalty0 (8):\penalty0
  443--455, 1992.
\newblock \doi{https://doi.org/10.1002/pat.1992.220030805}.
\newblock URL
  \url{https://onlinelibrary.wiley.com/doi/abs/10.1002/pat.1992.220030805}.

\bibitem[py~spy(2022)]{py-spy}
py~spy.
\newblock "\url{https://github.com/benfred/py-spy}", 2022.

\bibitem[Rdkit(2022)]{rdkit}
Rdkit.
\newblock "\url{http://www.rdkit.org/}",
  "\url{https://github.com/rdkit/rdkit}", 2022.

\bibitem[Riniker and Landrum(2015)]{torsion_angle}
S.~Riniker and G.~A. Landrum.
\newblock Better informed distance geometry: using what we know to improve
  conformation generation.
\newblock \emph{Journal of chemical information and modeling}, 55\penalty0
  (12):\penalty0 2562--2574, 2015.

\bibitem[Rogers and Hahn(2010)]{morganfp}
D.~Rogers and M.~Hahn.
\newblock Extended-connectivity fingerprints.
\newblock \emph{Journal of chemical information and modeling}, 50\penalty0
  (5):\penalty0 742--754, 2010.

\bibitem[Shi et~al.(2020)Shi, Xu, Zhu, Zhang, Zhang, and Tang]{graphaf}
C.~Shi, M.~Xu, Z.~Zhu, W.~Zhang, M.~Zhang, and J.~Tang.
\newblock Graphaf: a flow-based autoregressive model for molecular graph
  generation, 2020.

\bibitem[St.~John et~al.(2020)St.~John, Guan, Kim, Kim, and Paton]{alfabet}
P.~C. St.~John, Y.~Guan, Y.~Kim, S.~Kim, and R.~S. Paton.
\newblock Prediction of organic homolytic bond dissociation enthalpies at near
  chemical accuracy with sub-second computational cost.
\newblock \emph{Nature communications}, 11\penalty0 (1):\penalty0 2328, 2020.

\bibitem[Ståhl et~al.(2019)Ståhl, Falkman, Karlsson, Mathiason, and
  Boström]{rl_mol_3}
N.~Ståhl, G.~Falkman, A.~Karlsson, G.~Mathiason, and J.~Boström.
\newblock Deep reinforcement learning for multiparameter optimization in de
  novo drug design.
\newblock \emph{Journal of Chemical Information and Modeling}, 59\penalty0
  (7):\penalty0 3166--3176, 2019.
\newblock \doi{10.1021/acs.jcim.9b00325}.
\newblock URL \url{https://doi.org/10.1021/acs.jcim.9b00325}.
\newblock PMID: 31273995.

\bibitem[Sutton and Barto(2018)]{intro_rl}
R.~S. Sutton and A.~G. Barto.
\newblock \emph{Reinforcement Learning: An Introduction}.
\newblock The MIT Press, second edition, 2018.
\newblock URL \url{http://incompleteideas.net/book/the-book-2nd.html}.

\bibitem[Vakarelska-Popovska and Velkov(2016)]{vakarelska2016monohydroxy}
M.~H. Vakarelska-Popovska and Z.~Velkov.
\newblock Monohydroxy flavones. part iv: Ehthalpies of different ways of o--h
  bond dissociation.
\newblock \emph{Computational and Theoretical Chemistry}, 1077:\penalty0
  87--91, 2016.

\bibitem[Varatharajan and Pushparani(2018)]{anti_application_1}
K.~Varatharajan and D.~Pushparani.
\newblock Screening of antioxidant additives for biodiesel fuels.
\newblock \emph{Renewable and Sustainable Energy Reviews}, 82:\penalty0
  2017--2028, 2018.
\newblock ISSN 1364-0321.
\newblock \doi{https://doi.org/10.1016/j.rser.2017.07.020}.
\newblock URL
  \url{https://www.sciencedirect.com/science/article/pii/S1364032117310870}.

\bibitem[You et~al.(2018)You, Liu, Ying, Pande, and Leskovec]{gcpn}
J.~You, B.~Liu, Z.~Ying, V.~Pande, and J.~Leskovec.
\newblock Graph convolutional policy network for goal-directed molecular graph
  generation.
\newblock In S.~Bengio, H.~Wallach, H.~Larochelle, K.~Grauman, N.~Cesa-Bianchi,
  and R.~Garnett, editors, \emph{Advances in Neural Information Processing
  Systems}, volume~31. Curran Associates, Inc., 2018.
\newblock URL
  \url{https://proceedings.neurips.cc/paper_files/paper/2018/file/d60678e8f2ba9c540798ebbde31177e8-Paper.pdf}.

\bibitem[Zhang and Wang(2002)]{anti_bde_2}
H.-Y. Zhang and L.-F. Wang.
\newblock Theoretical elucidation on structure–antioxidant activity
  relationships for indolinonic hydroxylamines.
\newblock \emph{Bioorganic \& Medicinal Chemistry Letters}, 12\penalty0
  (2):\penalty0 225--227, 2002.
\newblock ISSN 0960-894X.
\newblock \doi{https://doi.org/10.1016/S0960-894X(01)00724-7}.
\newblock URL
  \url{https://www.sciencedirect.com/science/article/pii/S0960894X01007247}.

\bibitem[Zhang et~al.(2001)Zhang, Sun, and Chen]{bde_dft}
H.-Y. Zhang, Y.-M. Sun, and D.-Z. Chen.
\newblock O--h bond dissociation energies of phenolic compounds are determined
  by field/inductive effect or resonance effect? a dft study and its
  implication.
\newblock \emph{Quantitative Structure-Activity Relationships}, 20\penalty0
  (2):\penalty0 148--152, 2001.

\bibitem[Zhou et~al.(2019)Zhou, Kearnes, Li, Zare, and Riley]{moldqn}
Z.~Zhou, S.~Kearnes, L.~Li, R.~N. Zare, and P.~Riley.
\newblock Optimization of molecules via deep reinforcement learning.
\newblock \emph{Scientific reports}, 9\penalty0 (1):\penalty0 1--10, 2019.

\bibitem[Ziegler et~al.(2019)Ziegler, Stiennon, Wu, Brown, Radford, Amodei,
  Christiano, and Irving]{fine_tune_1}
D.~M. Ziegler, N.~Stiennon, J.~Wu, T.~B. Brown, A.~Radford, D.~Amodei,
  P.~Christiano, and G.~Irving.
\newblock Fine-tuning language models from human preferences.
\newblock \emph{arXiv preprint arXiv:1909.08593}, 2019.

\bibitem[Zubatyuk et~al.(2021)Zubatyuk, Smith, Nebgen, Tretiak, and
  Isayev]{AIMNet-nse}
R.~Zubatyuk, J.~S. Smith, B.~T. Nebgen, S.~Tretiak, and O.~Isayev.
\newblock Teaching a neural network to attach and detach electrons from
  molecules.
\newblock \emph{Nature Communications}, 12\penalty0 (1):\penalty0 4870, 2021.

\end{thebibliography}

\newpage
\appendix
\renewcommand{\thesection}{\Alph{section}.\arabic{section}}
\setcounter{section}{0}

\begin{appendices}

% \section{Protection of O-H Bond}\label{app:protection_of_oh}
\section{Examples of Action Molecules}\label{app:protection_of_oh}

% Since the BDE property is the lowest BDE among all O-H bonds,
% %
% an implicit restriction is that the generated molecules must have at least one O-H bond, 
% otherwise, the BDE properties are undefined and the molecules are not need.
% %
% This restriction is not guaranteed in the MolDQN and MT-MolDQN, 
% %
% because the algorithms modify molecules by adding an atom, adding a bond, or removing a bond. 
% %
% The O-H bond may be accidentally broken by these modifications (e.g. Step 2 in Figure~\ref{fig:actions}), 
% with the result that the molecule, and all the molecules that follow it, are invalid.
% %
% Although it is possible that the molecule will repair the bond or have another O-H bond in the next modification steps (e.g. Step 3 in Figure~\ref{fig:actions}), 
% molecules are still forced to have at least one O-H bond after every modification and will use the lowest BDE of them. 
% %
% This approach removes only a few invalid modifications from over a hundred action molecules, which has a trivial impact on the exploration of molecular space.

\begin{figure}[!h]
  \includegraphics[trim=0 0 0 0,clip,width=\textwidth]{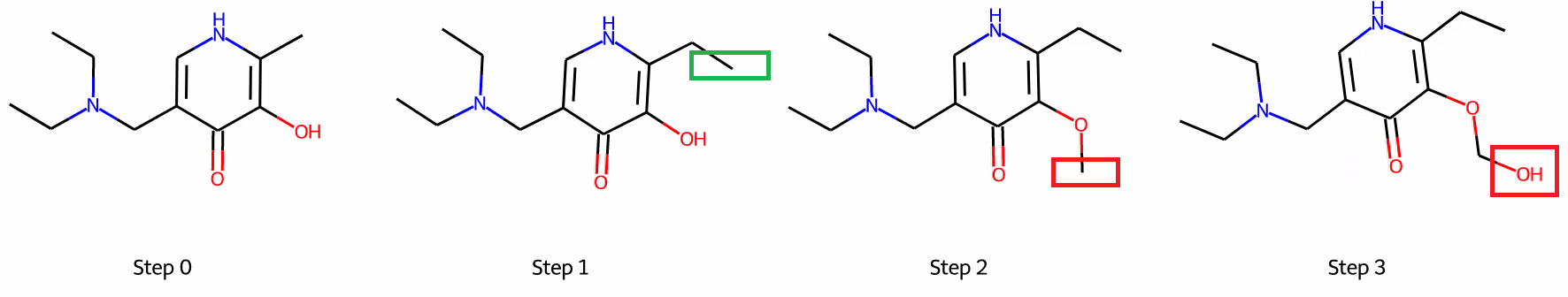}\hfil
\caption{Examples of valid modification (green rectangle) and invalid modifications (red rectangles). Each molecule has added an atom to the previous molecule on the left. The modifications that break the last O-H bond in step 2 are invalid. The atom addition in step 3 is also invalid because it happens after the invalid atom addition in step 2.}
\label{fig:actions} 
\end{figure}

% \newpage
% \section{Avoiding Invalid 3D Conformers}\label{app:avoiding_invalid_3d_conformers}
\section{Examples of Invalid 3D conformers \& Results of Avoiding Method}\label{app:avoiding_invalid_3d_conformers}

% AIMNet-NSE uses the 3D conformers of the molecules as its input, while the Alfabet uses SMILES strings. 
% For AIMNet-NSE, RDKit \cite{rdkit} is used to calculate the 3D coordinates of all atoms in the molecules \cite{embedMolecule1}. 
% But some generated molecules may not have a valid 3D conformer,
% even if they have a valid SMILES string and a molecule graph.
% %
% Because the MolDQN algorithm only takes care of the valence of atoms, while 3D conformers have much more chemical restrictions and special rules \cite{embed_uff}, such as the torsion angles \cite{torsion_angle} and aromatic ring.
% %
% An example molecule that has an invalid 3D conformer is shown in Figure~\ref{fig:counts_invalid_3d}.
% %
% The MolDQN has done some work on this, such as limiting the size of new rings, but their efforts are still far from being complete. 
% %
% In order to avoid generating molecules without 
% any valid 3D conformations, 
% RL agent is required to learn how to avoid these invalid conformers, rather than employing a huge number of human handwritten chemistry rules.
% %
% The reward of invalid molecules is set to -1000,
% which is much less than the normal rewards (0.8-2.5).
% % the experiment results do show that the RL agent has learned to avoid these obstacles in Section~\ref{sec:result}.
% Figure~\ref{fig:counts_invalid_3d} also traced how many invalid 3D conformers are generated in training episodes, 
% and the results show that RL agent has learned to avoid these obstacles.
\begin{figure}[h]
    \centering
    \begin{subfigure}{}
      \includegraphics[trim=0 0 0 0,clip,width=0.3\textwidth]{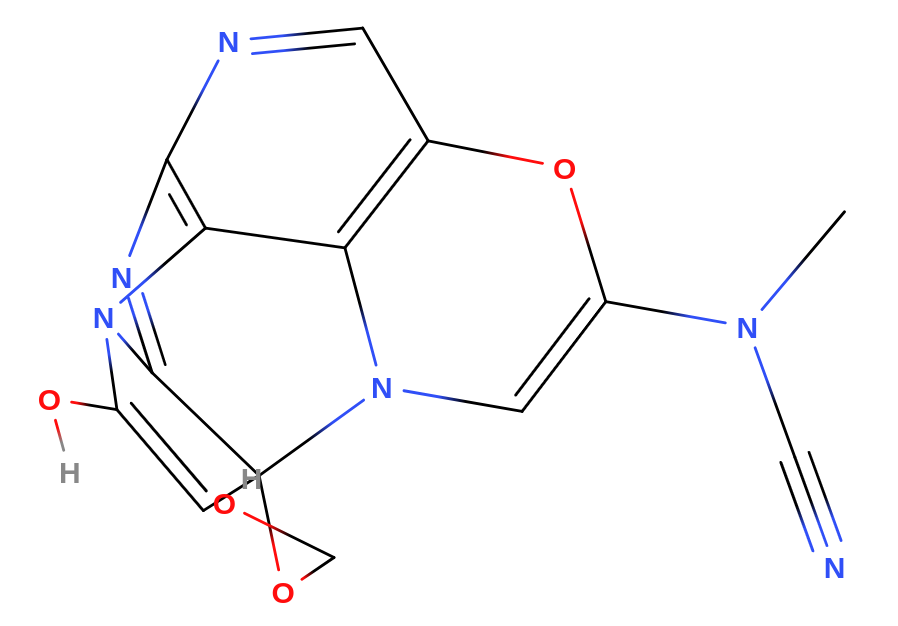}\hfil
    \end{subfigure}
    \begin{subfigure}{}
      \includegraphics[trim=0 0 0 0,clip,width=0.3\textwidth]{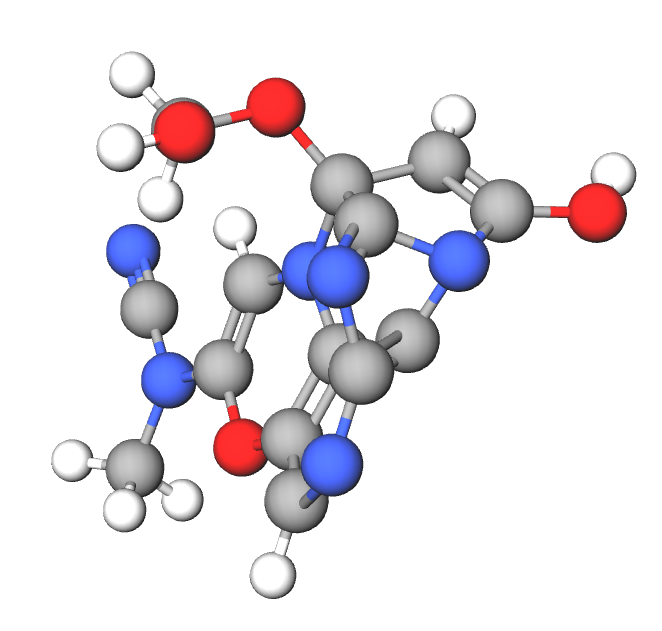}\hfil
    \end{subfigure}
    \begin{subfigure}{}
      \includegraphics[trim=0 0 0 0,clip,width=0.3\textwidth]{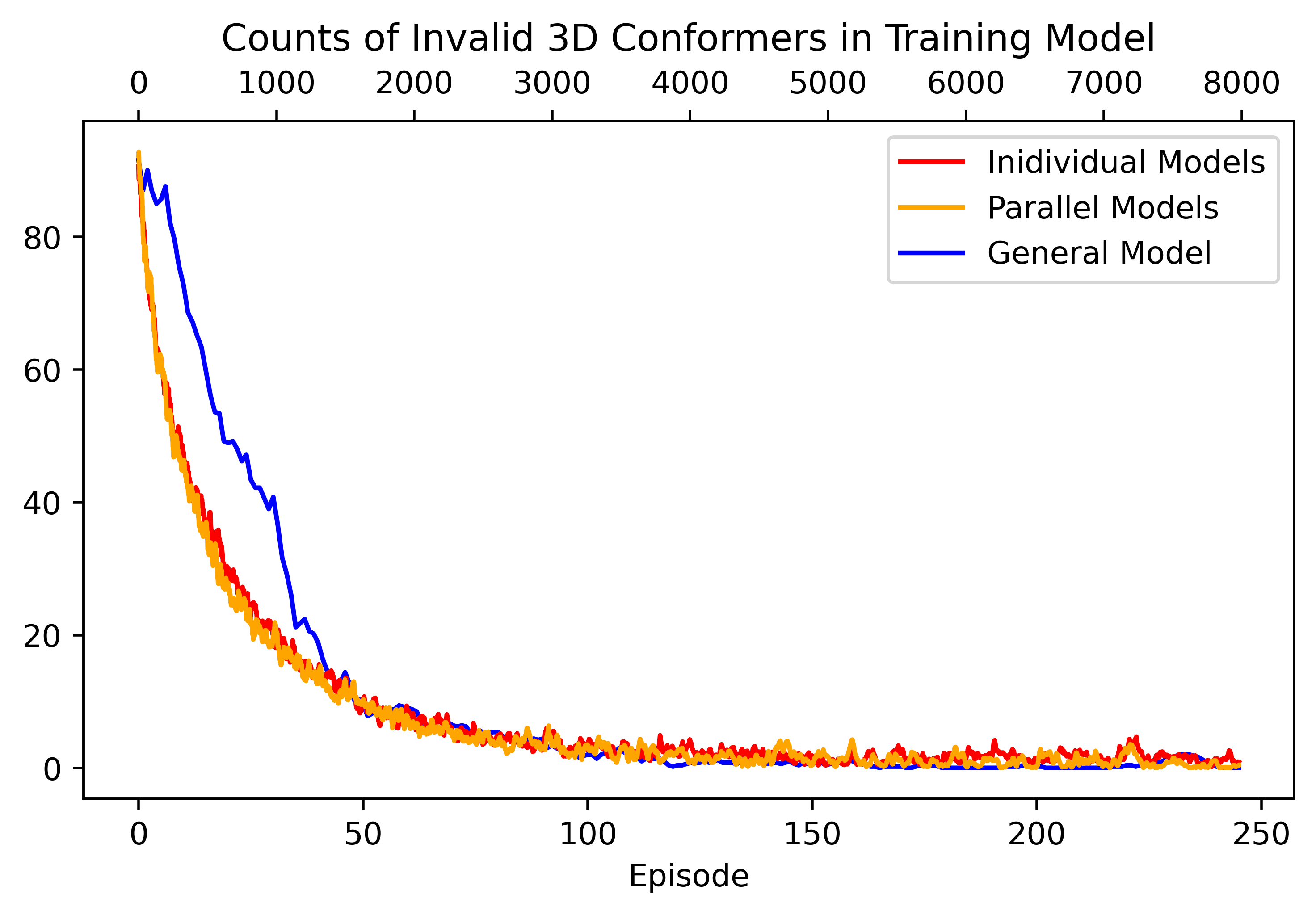}\hfil
    \end{subfigure}
    \caption{
    Left: The 2D representation of the generated molecule with invalid 3D conformers.
    Middle:  An example of the invalid 3D conformer.
    Right:  The RL agent learns to avoid invalid 3D conformers for individual models, parallel models, and general model.
    The individual and parallel models were trained with 8000 episodes and the general model was trained with 250 episodes.
    }
    \label{fig:counts_invalid_3d}
\end{figure}
\newpage
% \section{\textcolor{red}{Experiments Parameters}}\label{app:expr_para}
\section{Experiments Parameters}\label{app:expr_para}

\begin{table}[ht]
\centering
\begin{tabular}{|c|c|c|c|c|}
\hline
Model      & Initial Epsilon & Epsilon Decay & Max Training Batch Size & Starter  \\ \hline
Individual & 1.0             & 0.999         & 128                     & Torchrun \\ \hline
Parallel   & 1.0             & 0.999         & 128                     & Torchrun \\ \hline
General    & 1.0             & 0.970         & 512                     & Slurm    \\ \hline
Fine-Tuned & 0.5             & 0.961         & 128                     & Torchrun \\ \hline
\end{tabular}
\caption{This table summarizes additional experiment parameters that are \textbf{different} among models.}
\end{table}

\begin{table}[ht]
\centering
\begin{tabular}{|c|c|c|c|}
\hline
Max Steps/Episodes           & 10   & Allowed Atoms      & C, O, N \\ \hline
Update Episodes              & 1    & Allowed Rings      & 3, 5, 6 \\ \hline
DDP Backend                  & gloo & Fingerprint Radius & 3       \\ \hline
Replay Buffer Size           & 4000 & Fingerprint Length & 2048    \\ \hline
BDE Weight                   & 0.8  & BDE Factor         & 0.9     \\ \hline
IP Weight                    & 0.2  & IP Factor          & 0.8     \\ \hline
$\gamma$ Weight & 0.5          & Optimizer          & Adam    \\ \hline
Discount Factor              & 1.0  & Learning Rate      & 1e-4    \\ \hline
\end{tabular}
\caption{This table summarizes the experiment parameters that are the \textbf{same} among models.}
\end{table}

% \section{\textcolor{red}{Comparison with Related Works}}\label{app:compare_with_gnn}
\section{Comparison with Related Works}\label{app:compare_with_gnn}

\begin{table}[ht]\label{table:top3_gms}
\centering
\begin{tabular}{|c|ccc|ccc|}
\hline
          & \multicolumn{3}{c|}{QED}                                        & \multicolumn{3}{c|}{Penalized LogP}                             \\ \hline
          & \multicolumn{1}{c|}{1st}   & \multicolumn{1}{c|}{2nd}   & 3rd   & \multicolumn{1}{c|}{1st}   & \multicolumn{1}{c|}{2nd}   & 3rd   \\ \hline
MolDQN    & \multicolumn{1}{c|}{0.948} & \multicolumn{1}{c|}{0.944} & 0.943 & \multicolumn{1}{c|}{11.84} & \multicolumn{1}{c|}{11.84} & 11.82 \\ \hline
DA-MolDQN & \multicolumn{1}{c|}{0.948} & \multicolumn{1}{c|}{0.948} & 0.947 & \multicolumn{1}{c|}{7.12}  & \multicolumn{1}{c|}{7.07}  & 6.94  \\ \hline
GCPN      & \multicolumn{1}{c|}{0.948} & \multicolumn{1}{c|}{0.948} & 0.947 & \multicolumn{1}{c|}{6.56}  & \multicolumn{1}{c|}{6.46}  & 6.40  \\ \hline
GraphAF   & \multicolumn{1}{c|}{0.948} & \multicolumn{1}{c|}{0.947} & 0.947 & \multicolumn{1}{c|}{5.63}  & \multicolumn{1}{c|}{5.60}  & 5.44  \\ \hline
\end{tabular}
% \caption{Top 3 Generated Molecules}
\caption{The top 3 molecules are collected from both training and testing. 
The QED properties of top molecules are similar among the 4 models.
The Plogp of MolDQN significantly outperforms DA-MolDQN and other algorithms because the PlogP is maximized by simply adding carbon atoms. 
% Resulting that 
As a result,
the generated molecules are obviously not drug-like~\cite{moldqn}.
}
\end{table}

\begin{figure}[!h]\label{fig:compare_qed_plogp}
    \centering
    \begin{subfigure}{}
        \includegraphics[trim=0 0 0 0,clip,width=0.47\textwidth]{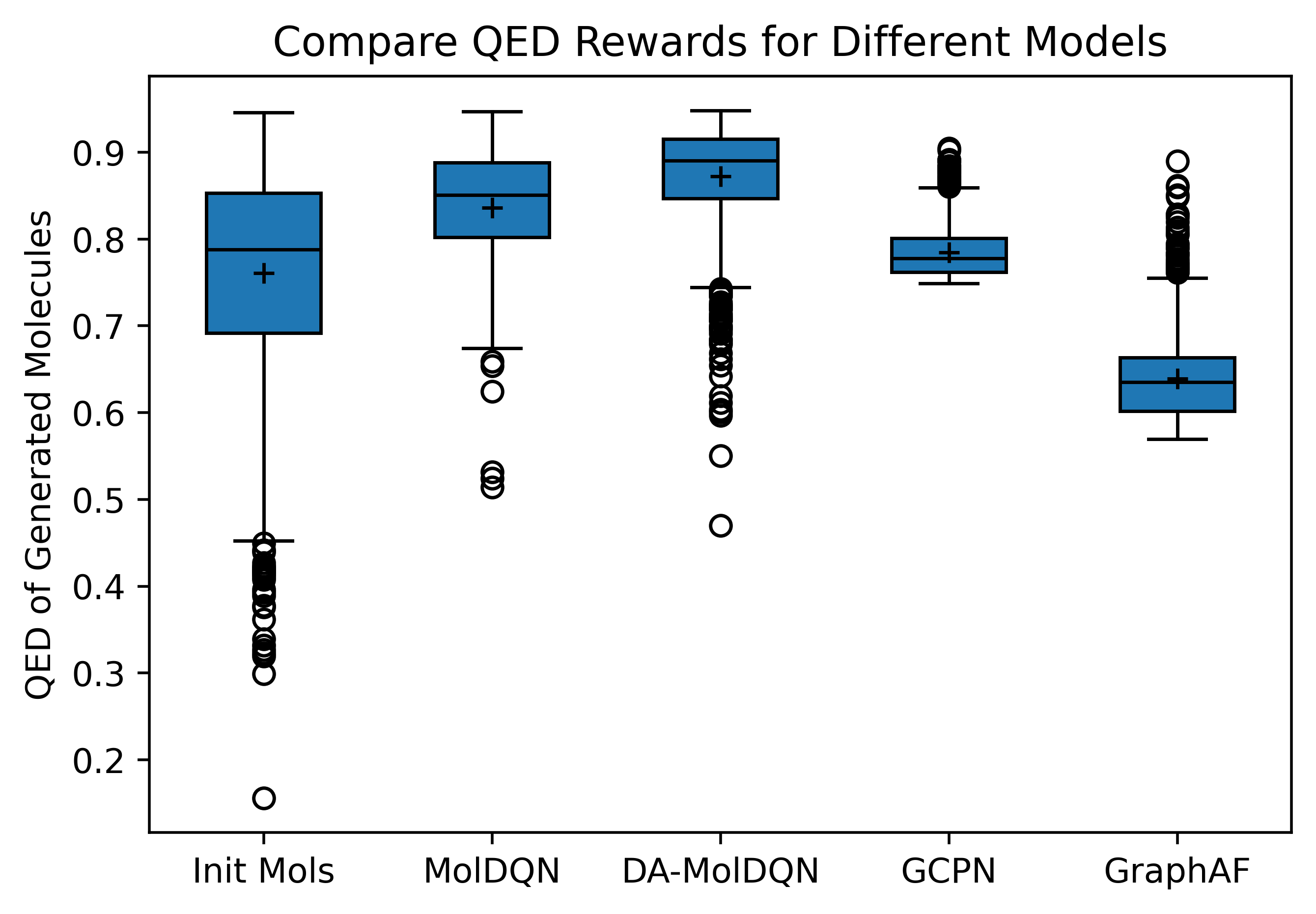}
    \end{subfigure}
    \begin{subfigure}{}
        \includegraphics[trim=0 0 0 0,clip,width=0.47\textwidth]{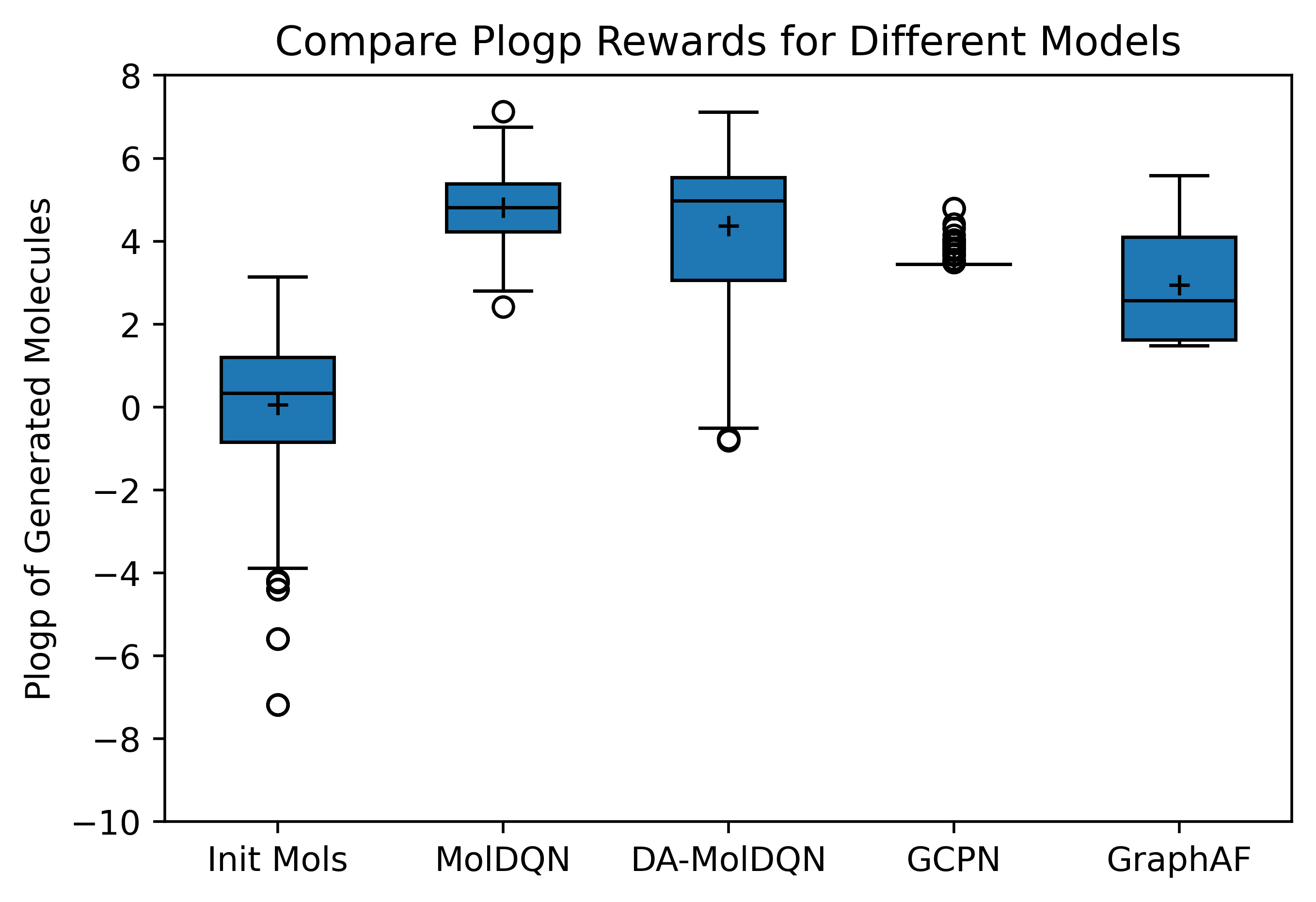}
    \end{subfigure}
    
    \caption{The figure shows the molecule QED and PlogP rewards for MolDQN, DA-MolDQN, GCPN, and GraphAF.
    256 molecules are optimized by MolDQN, while 1024 molecules are generated by the DA-MolDQN fine-tuned models. For GCPN and GraphAF, the molecules are generated in the same wall clock time and only show the top 1024 molecules.
    }
\end{figure}

% generated mols:
%         QED     plogp
% MolDQN  256     256
% DA-M    1024    1024
% GCPN    47000   18000
% GraphAF 6400    6400

In the comparison experiments, the QED and penalized logP (PlogP) rewards were tested. The results were compared with GCPN \cite{gcpn} and GraphAF \cite{graphaf}, which are state-of-the-art molecule generation algorithms based on GNN.
% and are integrated into TorchDrug.
% Their TorchDrug  
All models are trained with the Zinc250k data set. 256 and 1024 molecules are randomly selected to train for MolDQN and DA-MolDQN. The training parameters are the same as antioxidant experiments. GCPN and GraphAF are trained with the whole Zinc250k data set. The models are trained with (1-10) epochs, and the best results are present. 
% 256 molecules are optimized by MolDQN, while 1024 or more molecules are generated by DA-MolDQN, GCPN, and GraphAF.
The results % in Table~\ref{table:top3_gms} and Figure~\ref{fig:compare_qed_plogp} 
indicate that our DA-MolDQN achieves similar performance in optimizing top molecules, generates molecules with higher QEDs and avoids failures in PlogP optimization. 

% \paragraph{Top Generated Molecules}
% The top 3 generated molecules are shown in Table~\ref{table:top3_gms}. These molecules are collected from both training and testing. 
% The QED properties of top molecules are similar among the 4 models.
% The Plogp of MolDQN significantly outperforms DA-MolDQN and other algorithms because the PlogP is maximized by simply adding carbon atoms. Resulting that the generated molecules are obviously not drug-like~\cite{moldqn}.
% These results indicate that our DA-MolDQN achieves similar performance in optimizing top molecules and avoids failures in PlogP optimization.

% \paragraph{Molecules Generated by Trained Models}
% The molecules generated by trained models are presented in Figure~\ref{fig:compare_qed_plogp}. 
% We present the 1024 molecules generated by MolDQN and DA-MolDQN. 
% The results show that our DA-MolDQN outperforms the MolDQN, GCNP, and GraphAF.

% MolDQN: This is partly because maximizing logP corresponds to a simple policy: adding carbon atoms wherever possible. This straightforward policy does not require much exploration effort, and can be regarded as a greedy policy

% \paragraph{QED Optimization Results}

% The range of QED is $\left[0, 1\right]$. 

% \newpage
\section{More Examples of Proposed Molecules \& Generating Path}\label{app:public_gm_all}
\begin{figure}[!h] % add ht
    \centering
    \begin{subfigure}{}
      \includegraphics[trim=0 0 0 0,clip,width=0.15\textwidth]{Figures/generate_path/trial_673200_3_0.png}\hfil
    \end{subfigure}
    \begin{subfigure}{}
      \includegraphics[trim=0 0 0 0,clip,width=0.15\textwidth]{Figures/generate_path/trial_673200_3_1.png}\hfil
    \end{subfigure}
    \begin{subfigure}{}
      \includegraphics[trim=0 0 0 0,clip,width=0.15\textwidth]{Figures/generate_path/trial_673200_3_2.png}\hfil
    \end{subfigure}
    \begin{subfigure}{}
      \includegraphics[trim=0 0 0 0,clip,width=0.15\textwidth]{Figures/generate_path/trial_673200_3_3.png}\hfil
    \end{subfigure}
    \begin{subfigure}{}
      \includegraphics[trim=0 0 0 0,clip,width=0.15\textwidth]{Figures/generate_path/trial_673200_3_4.png}\hfil
    \end{subfigure}
    \begin{subfigure}{}
      \includegraphics[trim=0 0 0 0,clip,width=0.15\textwidth]{Figures/generate_path/trial_673200_3_5.png}\hfil
    \end{subfigure}

    \begin{subfigure}{}
      \includegraphics[trim=0 0 0 0,clip,width=0.15\textwidth]{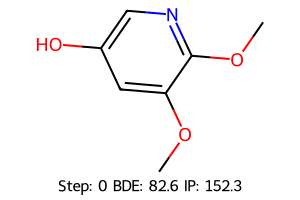}\hfil
    \end{subfigure}
    \begin{subfigure}{}
      \includegraphics[trim=0 0 0 0,clip,width=0.15\textwidth]{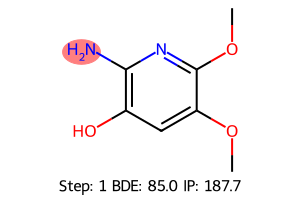}\hfil
    \end{subfigure}
    \begin{subfigure}{}
      \includegraphics[trim=0 0 0 0,clip,width=0.15\textwidth]{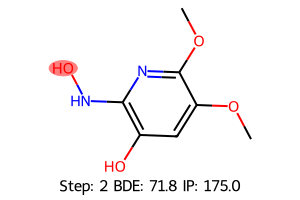}\hfil
    \end{subfigure}
    \begin{subfigure}{}
      \includegraphics[trim=0 0 0 0,clip,width=0.15\textwidth]{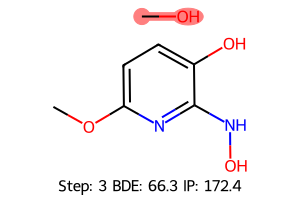}\hfil
    \end{subfigure}
    \begin{subfigure}{}
      \includegraphics[trim=0 0 0 0,clip,width=0.15\textwidth]{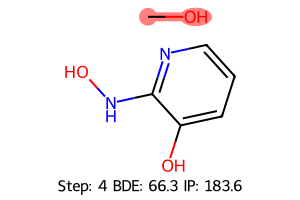}\hfil
    \end{subfigure}
    \begin{subfigure}{}
      \includegraphics[trim=0 0 0 0,clip,width=0.15\textwidth]{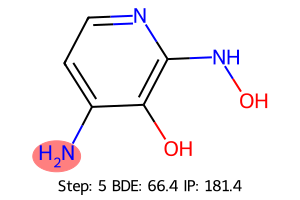}\hfil
    \end{subfigure}

    \begin{subfigure}{}
      \includegraphics[trim=0 0 0 0,clip,width=0.15\textwidth]{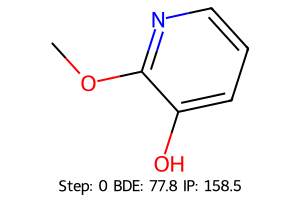}\hfil
    \end{subfigure}
    \begin{subfigure}{}
      \includegraphics[trim=0 0 0 0,clip,width=0.15\textwidth]{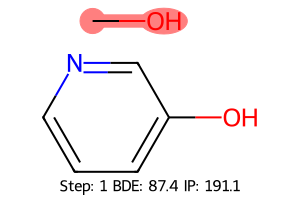}\hfil
    \end{subfigure}
    \begin{subfigure}{}
      \includegraphics[trim=0 0 0 0,clip,width=0.15\textwidth]{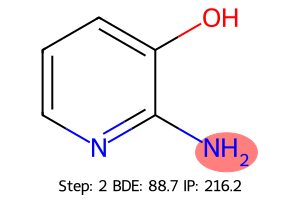}\hfil
    \end{subfigure}
    \begin{subfigure}{}
      \includegraphics[trim=0 0 0 0,clip,width=0.15\textwidth]{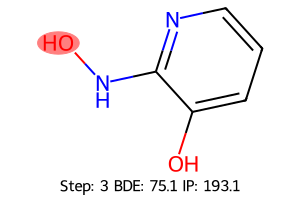}\hfil
    \end{subfigure}
    \begin{subfigure}{}
      \includegraphics[trim=0 0 0 0,clip,width=0.15\textwidth]{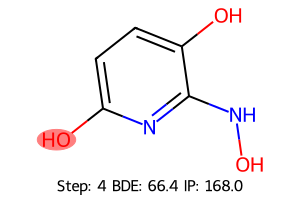}\hfil
    \end{subfigure}
    \begin{subfigure}{}
      \includegraphics[trim=0 0 0 0,clip,width=0.15\textwidth]{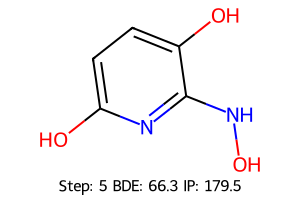}\hfil
    \end{subfigure}

    \begin{subfigure}{}
      \includegraphics[trim=0 0 0 0,clip,width=0.15\textwidth]{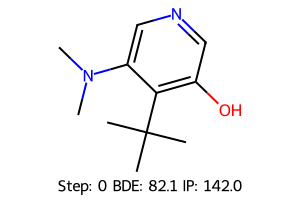}\hfil
    \end{subfigure}
    \begin{subfigure}{}
      \includegraphics[trim=0 0 0 0,clip,width=0.15\textwidth]{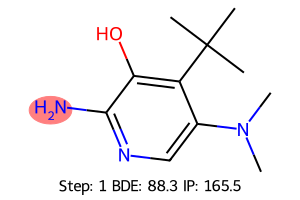}\hfil
    \end{subfigure}
    \begin{subfigure}{}
      \includegraphics[trim=0 0 0 0,clip,width=0.15\textwidth]{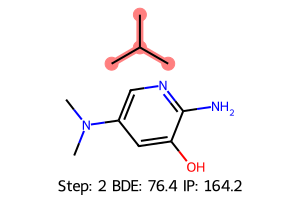}\hfil
    \end{subfigure}
    \begin{subfigure}{}
      \includegraphics[trim=0 0 0 0,clip,width=0.15\textwidth]{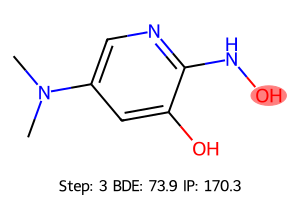}\hfil
    \end{subfigure}
    \begin{subfigure}{}
      \includegraphics[trim=0 0 0 0,clip,width=0.15\textwidth]{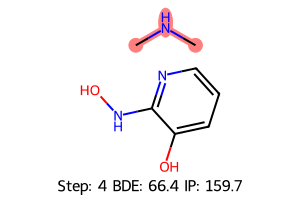}\hfil
    \end{subfigure}
    \begin{subfigure}{}
      \includegraphics[trim=0 0 0 0,clip,width=0.15\textwidth]{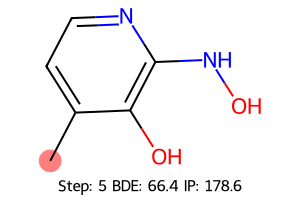}\hfil
    \end{subfigure}

    \begin{subfigure}{}
      \includegraphics[trim=0 0 0 0,clip,width=0.15\textwidth]{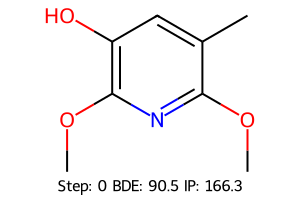}\hfil
    \end{subfigure}
    \begin{subfigure}{}
      \includegraphics[trim=0 0 0 0,clip,width=0.15\textwidth]{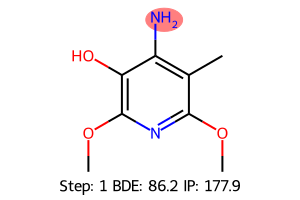}\hfil
    \end{subfigure}
    \begin{subfigure}{}
      \includegraphics[trim=0 0 0 0,clip,width=0.15\textwidth]{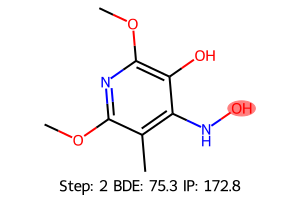}\hfil
    \end{subfigure}
    \begin{subfigure}{}
      \includegraphics[trim=0 0 0 0,clip,width=0.15\textwidth]{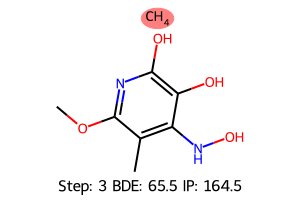}\hfil
    \end{subfigure}
    \begin{subfigure}{}
      \includegraphics[trim=0 0 0 0,clip,width=0.15\textwidth]{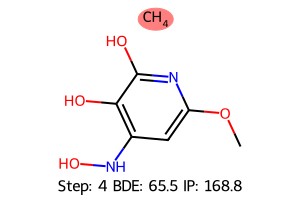}\hfil
    \end{subfigure}
    \begin{subfigure}{}
      \includegraphics[trim=0 0 0 0,clip,width=0.15\textwidth]{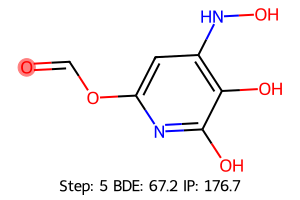}\hfil
    \end{subfigure}

    \caption{The figure shows several optimized antioxidants and their generating paths.
    The molecules in Step 0 are the initial molecules and the optimized molecules are in Step 5.
    The modifications are highlighted and the unconnected atoms are removed from the current molecules. 
    }
    \label{fig:gm_public_all}
\end{figure}

% \section{to do}\label{app:not_helpful_optimizations}
% This problem will not be solved by increasing the maximum steps of an episode or by adding more episodes to the training model.
\section{Result of DFT Validation}\label{app:dft_validation}
% 7 proposed molecules are selected from the antioxidants optimized by the parallel models for evaluation with DFT simulations.
% DFT simulation result of the general model is underway.
% %The molecules were simulated with DFT to calculate the energy required for antioxidant ionization. 
% %The ionization split the molecule into radicals. The energy state of resulting molecules and radicals was evaluated.
% The table below shows the comparison of predicted and DFT results, for BDE and IP.
% The results indicate that the error is within tolerance for optimizing these molecules.
\begin{table}[ht]
\begin{tabular}{|c|c|c|c|c|c|c|c|c|}
\hline
No. & Initial BDE & BDE\textsubscript{ML}  & BDE\textsubscript{DFT}   & Initial IP & IP\textsubscript{ML}    & IP\textsubscript{DFT}     & Similarity & SA Score \\ \hline
1   & 76.9        & 72.4 & 77.91 & 139.2      & 164.6 & 168.4  & 0.13       & 2.49     \\ \hline
2   & 76.9        & 74.4 & 76.04 & 139.2      & 155.9 & 147.87 & 0.12       & 2.53     \\ \hline
3   & 67.1        & 75.7 & 81.19 & 113.8      & 174.3 & 187.99 & 0.18       & 2.83     \\ \hline
4   & 67.1        & 73.0 & 77.82 & 113.8      & 166.9 & 163.86 & 0.18       & 2.80     \\ \hline
5   & 64.3        & 72.5 & 77.61 & 113.4      & 161.5 & 166.05 & 0.19       & 2.79     \\ \hline
6   & 64.3        & 69.7 & 74.85 & 113.4      & 156.8 & 150.65 & 0.18       & 2.88     \\ \hline
7   & 64.3        & 67,8 & 71.53 & 113.4      & 158.1 & 141.45 & 0.19       & 2.52     \\ \hline
\end{tabular}
\caption{The table shows the DFT validation results of new antioxidants generated by parallel models.
The initial BDE and IP are generated by DFT.
BDE\textsubscript{ML} and IP\textsubscript{ML} are predicted by the property predictors.
% to do: add unit: kcal/mol
}
\end{table}

\end{appendices}

\end{document}